\documentclass{article}

\usepackage[square,numbers]{natbib}
\usepackage[preprint]{neurips_2026}

\usepackage[utf8]{inputenc} % allow utf-8 input
\usepackage[T1]{fontenc}    % use 8-bit T1 fonts
\usepackage{hyperref}       % hyperlinks
\usepackage{url}            % simple URL typesetting
\usepackage{booktabs}       % professional-quality tables
\usepackage{amsfonts}       % blackboard math symbols
\usepackage{nicefrac}       % compact symbols for 1/2, etc.
\usepackage{microtype}      % microtypography
\usepackage{xcolor}         % colors
\usepackage{graphicx}
\usepackage{subcaption}
\usepackage{amsmath}
\usepackage{algorithm}
\usepackage{algpseudocode}
\usepackage{multirow} 
\usepackage{bm}
\usepackage{tcolorbox}
\usepackage{adjustbox}
\usepackage{caption}
\usepackage{listings}
\usepackage{xspace}

\usepackage{wrapfig}
\usepackage{subfiles}
\usepackage{bbm}

\newcommand{\pihl}{\pi_{\mathrm{HL}}} %$\xspace}
\newcommand{\pill}{\pi_{\mathrm{LL}}} %$\xspace}
\newcommand{\herakles}{{\sc Herakles}\xspace} %$\xspace}

\title{HERAKLES: Hierarchical Skill Compilation \\ for Open-ended LLM Agents
}

\author{%
  Thomas Carta \\
  Inria(Flowers) \\
  University of Bordeaux, France \\
  \And 
  Clément Romac \\
  Inria(Flowers) \\
  University of Bordeaux, France \\
  Hugging Face \\
  \And
  Loris Gaven \\
  Inria(Flowers) \\
  University of Bordeaux, France \\
  \And
  Pierre-Yves Oudeyer \\
  Inria(Flowers) \\
  University of Bordeaux, France \\
  \And
  Olivier Sigaud \\
  Sorbonne Université (ISIR) \\
  Paris, France \\
  \And
  Sylvain Lamprier \\
  Univ Angers (LERIA) \\
  SFR MATHSTIC, F-49000 \\
  Angers, France \\
}

\usepackage{xcolor}

\begin{document}

\maketitle

\begin{abstract}

We study goal-conditioned reinforcement learning in partially observable environments with sparse rewards and large, structured goal spaces. In such settings, complex goals often require composing simpler skills, but learning these compositions efficiently remains challenging. This difficulty is particularly relevant in open-ended exploration settings, where agents are exposed to increasingly complex goals over time and must continuously expand and reorganize their skill repertoire. We assume the goal space admits prerequisite relations, enabling latent decomposition of tasks into subgoals, and leverage language to represent and reason over these goals. To exploit this structure, we propose HERAKLES, a hierarchical agent that jointly learns a high-level LLM policy and a low-level controller. The high-level policy selects subgoals among those the low-level can reliably achieve, while the low-level executes them and progressively compiles successful behaviors into reusable skills. Both policies are trained concurrently: the high-level guides exploration and structures behavior, while the low-level distills trajectories into efficient goal-conditioned skills. As training progresses, more goals become directly executable, enabling scalable skill composition. This results in a hybrid system combining planning and execution, improving efficiency and adaptation in open-ended, compositional environments.

\end{abstract}

\section{Introduction}

%have produced agents with human-level performance in vision and language goals, largely due to foundation models trained on large-scale internet data~\cite{yu_cocacontrastivecaptionersimagetext2022, openai_gpt4technicalreport2024, deepseekai_deepseekr1incentivizingreasoningcapability2025}. However, the static and limited nature of these datasets raises concerns about their sufficiency for achieving truly general intelligence~\cite{villalobos_rundatalimitsllm2024}.
Recent breakthroughs in AI have enabled human-level competence on diverse behavioural tasks, largely due to foundation models trained on large-scale internet data~\cite{yu_cocacontrastivecaptionersimagetext2022, openai_gpt4technicalreport2024, deepseekai_deepseekr1incentivizingreasoningcapability2025}. However, the static and limited nature of these datasets and the models limits their use for life long growth in complexity~\cite{villalobos_rundatalimitsllm2024}. In contrast to these dataset-dependent approaches, humans acquire a remarkably diverse repertoire of skills continuously throughout life without relying on static datasets. One major objective of artificial intelligence is to endow agents with similar capabilities: learning without a dataset
through interactions with the environment and autonomously discovering and expanding a diverse skill set. The former is covered by RL methods while the latter, referred to as open-endedness, is considered fundamental for the development of 
life-long learning abilities of increasingly complex tasks that some have called Artificial General Intelligence (AGI) \cite{sigaud_definitionopenendedlearningproblems2024}, \cite{hughes_positionopenendedness2024}
%Artificial General Intelligence (AGI) \cite{sigaud_definitionopenendedlearningproblems2024} and even Artificial Super Intelligence (ASI) \cite{hughes_positionopenendedness2024}. 
According to \cite{colas_languageandculture2022}, such agents should be autotelic, meaning they are capable of generating, selecting and training on their own goals for continuous and open-ended self-improvement.

Recent approaches have begun leveraging foundation and generative models to construct open-ended and autotelic agents, as exemplified by MineDojo~\cite{fan_minedojo2022}, Voyager~\cite{wang_voyageropenendedembodiedagent2023}, OMNI \cite{zhangomni} or AutotelicLLM \cite{pourcel_autotelic2024}. These systems utilize large models to explore the goal space, either by autonomously generating novel goals and associated rewards or by structuring learning through goal prioritizaton over a pre-generated goal set~\cite{gaven_magellan2025}. However, as the goal space expands, increasingly complex goals emerge that require the composition of multiple actions. This combinatorial explosion in goal complexity, observed in~\cite{valmeekam_planningabilitieslargelanguage2023, adaptiveagentteam_humantimescaleadaptationopenendedtask2023, poesia_learning2024}, tends to impede the open-ended learning process. When each new goal demands an increasingly longer time to master, agent progress stalls.

%\paragraph{Hierarchical Learning}
Humans, as quintessential lifelong learners, face similar challenges in acquiring complex skills, such as mastering a new sport. However, they are always able to find new goals of increasing complexity that can be learned efficiently. %Nevertheless, they do not experience diminishing returns in learning. 
Building upon \cite{fitts_human1967}, Tsay et al. \cite{tsay_fundamental2024} demonstrate that humans employ hierarchical learning to overcome the complexity barrier: cognitive systems decompose new skills into simpler, previously internalized subskills handled by sensorimotor systems. As new goals are mastered, they are recursively encoded at lower levels, enabling rapid reuse in future learning.

This biological principle has inspired AI research. Hierarchical structures have been adapted to enable open-ended object recognition~\cite{kasaei_hierarchical2016} and Hierarchical Reinforcement Learning (HRL)~\cite{sutton_between1999, precup_temporal2000}. Several recent works, including \cite{hu_hierarchical2019, duminy_intrinsically2021}, \cite{ahuja_hierarchicalreinforcementlearningnatural2023}, and \cite{jiang_language2019}, have further incorporated natural language structures to facilitate goal decomposition. However, most of these methods assume predefined skills, often requiring pre-trained policies on these skills. This static setup is inherently incompatible with open-ended agents, which must continuously face novel and increasingly complex goals. 

In this work, we propose {\bf HERAKLES} for \textbf{H}i\textbf{ER}archic\textbf{A}l s\textbf{K}ill compi\textbf{L}ation for open \textbf{E}nded llm agent\textbf{S}. It is a method for open-ended autotelic agents that jointly learns high- and low-level policies in a complex goal space. \herakles extends prior HRL and language-based approaches to support continual adaptation without requiring pre-specified skills or extensive pre-training. Specifically, we train a high-level policy $\pihl$ to invoke a low-level policy $\pill$ on goals the latter has already mastered. Policy $\pihl$ is instantiated as a Large Language Model (LLM), leveraging its capacity to constrain exploration, select relevant skills, and operate effectively in a shifting skill landscape. Policy $\pihl$ is trained online over an automatic curriculum of increasingly difficult goals, while its learned skills are progressively distilled into $\pill$: a lightweight, computationally efficient model specialized in executing primitive actions.

We evaluate our method in the Crafter environment~\cite{hafner_benchmarking2022}, designed to assess a wide spectrum of agent capabilities within a unified open-ended framework.

\begin{figure}
\vspace{-\baselineskip}
    \centering
    \includegraphics[width=0.9\linewidth]{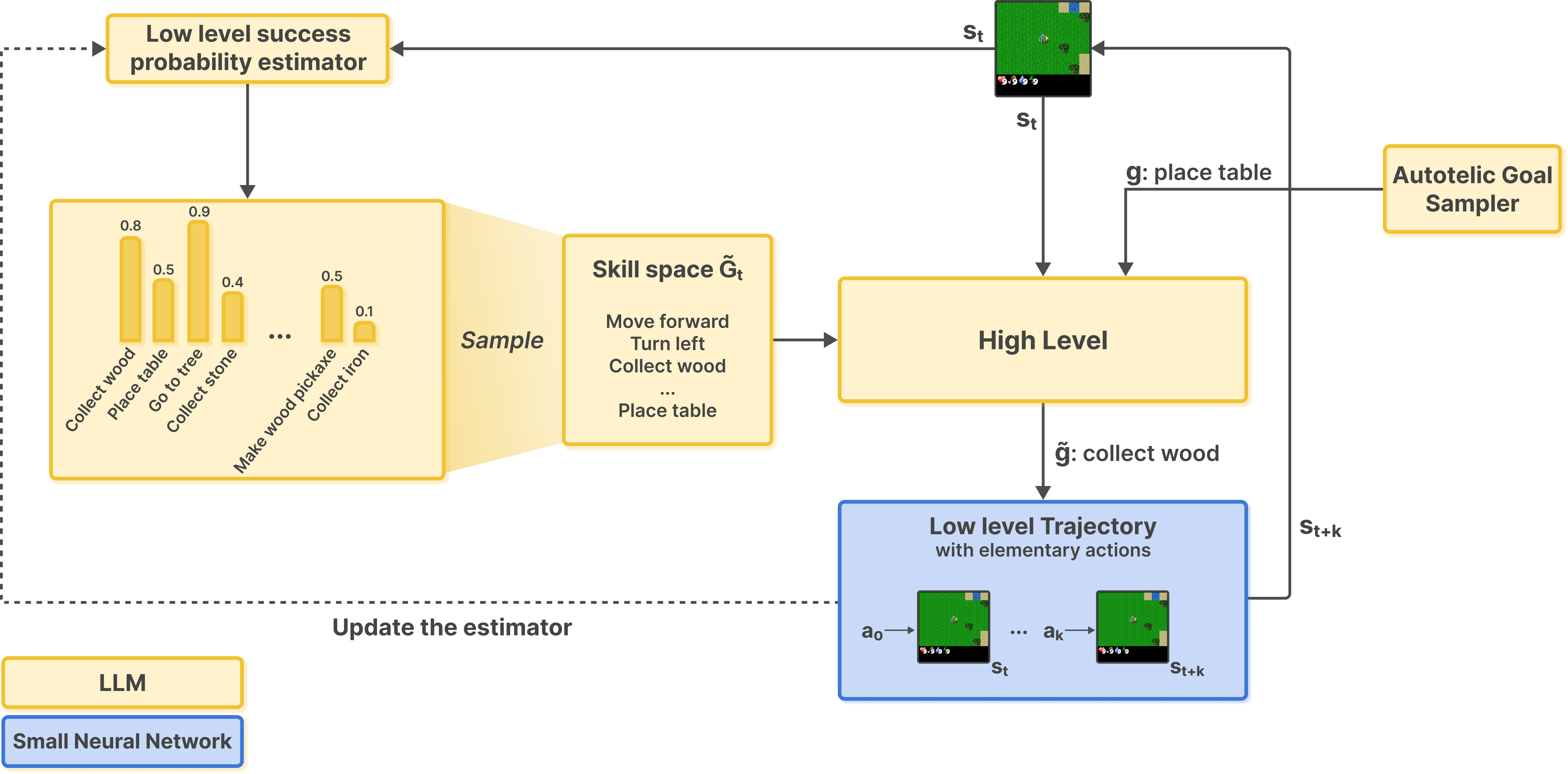}
    \caption{\textbf{HERAKLES} is a hierarchical RL agent for open-ended exploration, built on two policies. An LLM-based high-level policy iteratively selects goals for a low-level policy that solves them via elementary actions, restricting its selection to goals with high estimated success probability. Once \herakles masters a goal, the full trajectory of elementary actions is compiled into the low-level policy as a single new skill, which the high-level policy can then invoke directly.}
    \label{fig:inference}
\vspace{-\baselineskip}
\end{figure}

Our main research questions focus on the mechanism of effective learning, which is a key factor in the development of open-ended agents. They include:
\par\smallskip $\bullet$ How can we implement high- and low-level interaction and concurrent learning? How can we compile skills acquired by the hierarchical agent as a whole into $\pill$ to foster efficient learning?
%\par\smallskip $\bullet$ How does \herakles scale with the difficulty of the goal?
\par\smallskip $\bullet$ Is \herakles more sample-efficient than prior HRL approaches or LLM-only strategies?
\par\smallskip $\bullet$ Can \herakles generalize to novel goals without additional training?

%To answer these scientific questions, we describe how \herakles works in Section~\ref{methods} and test it in the Crafter environment \cite{hafner_benchmarking2022} in Section~\ref{experiments}.

\section{Related Work}

\paragraph{Neuroscience} 
Humans continuously acquire increasingly complex sensorimotor skills throughout their lifetime, a process driven by curiosity \cite{gottlieb_towards2018} and structured as a progression from 
deliberate reasoning to automatic retrieval~\cite{fitts_human1967, tsay_fundamental2024}.  This hierarchical organization is also reflected in neural architectures for perceptual learning \cite{riesenhuber_hierarchical1999}, and has motivated the use of language in  hierarchical developmental robotics, given its combinatorial and generalization  properties~\cite{cangelosi_integration2010}. Our work shares these conceptual foundations:  an LLM serves as a high-level policy over a low-level neural network, into which mastered  skills are continuously compiled, progressively expanding the subgoal space available for  learning more complex goals.

\paragraph{Hierarchical Reinforcement Learning and language}
Learning meaningful spatial and temporal abstractions to solve tasks more 
efficiently has long been studied in RL~\cite{dayan_feudal1992}. Building on the option framework~\cite{sutton_between1999, precup_temporal2000, bacon_option2017}, several deep HRL methods such as h-DQN~\cite{kulkarni_hierarchicaldeepreinforcementlearning2016}, \textsc{FuN}~\cite{vezhnevets_feudalnetworkshierarchicalreinforcement2017}, and HIRO~\cite{nachum_dataefficient2018} decompose goals into subgoals defined as points in a learned embedding space whose quality strongly conditions the hierarchy's performance~\cite{nachum_near2018}. \herakles decomposes goals into subgoals in the linguistic space leveraging the abilities of a pretrained LLM to compose skills \cite{ahn_icanisay2022, ahuja_hierarchicalreinforcementlearningnatural2023}. From a theoretical standpoint, the coupling between high- and low-level policies can be cast as a bilevel problem with a Stackelberg structure, for which two-timescale dynamics \cite{fiez_convergence2019} and RL-specific variants \cite{shen_principled2024, chakraborty_parl2024} provide convergence guarantees. Our training procedure is similarly motivated by the timescale separation between the LLM-based $\pihl$ and the lightweight $\pill$.%, though it does not implement the explicit hypergradient corrections or penalty terms these methods require.

%Language offers an alternative subgoal space whose compositional and generalization properties naturally fit the hierarchical structure of environments~\cite{kaplan_beating2017, luketina_surveyreinforcementlearninginformed2019, jiang_language2019}. LLMs have accordingly been used as high-level policies that decompose goals into skills executed by specialized low-level policies~\cite{ahn_icanisay2022, ahuja_hierarchicalreinforcementlearningnatural2023}, with VLM extensions to visual inputs~\cite{shi_hirobotopenendedinstruction2025, physical_pi2025}. However, all these methods assume that any goal can be decomposed into a small, expert-defined set of skills, and either pre-train $\pill$ on this fixed set~\cite{ahuja_hierarchicalreinforcementlearningnatural2023, shi_hirobotopenendedinstruction2025, physical_pi2025} or keep $\pihl$ frozen~\cite{ahn_icanisay2022}. Voyager~\cite{wang_voyageropenendedembodiedagent2023} similarly leverages an LLM to grow a skill library in an open-ended environment and keeps the LLM frozen. Our method also leverages language to structure the skill space, but neither relies on an expert-defined skill set nor pre-trains $\pill$: instead, previously mastered goals are reused as skills for reaching new ones.

Language provides an alternative subgoal space whose compositional and generalization properties match the hierarchical structure of environments~\cite{kaplan_beating2017, luketina_surveyreinforcementlearninginformed2019, jiang_language2019}. LLMs have thus been used as high-level policies decomposing goals into skills executed by specialized low-level policies~\cite{ahn_icanisay2022, ahuja_hierarchicalreinforcementlearningnatural2023}, including VLM extensions to visual inputs~\cite{shi_hirobotopenendedinstruction2025, physical_pi2025}. However, these methods assume goals decompose into a small expert-defined skill set, either used to pre-train $\pill$~\cite{ahuja_hierarchicalreinforcementlearningnatural2023, shi_hirobotopenendedinstruction2025, physical_pi2025} or with frozen $\pihl$~\cite{ahn_icanisay2022}. Similarly, Voyager~\cite{wang_voyageropenendedembodiedagent2023} grows a skill library while keeping the LLM frozen. Our method also structures skills through language, but neither relies on expert-defined skills nor pre-trains $\pill$: previously mastered goals are directly reused as skills for new ones.

\paragraph{Open-ended autotelic agent and language} Humans continuously face, and even invent, new goals throughout their lives \cite{gottlieb_towards2018}, an open-ended ability that has interested AI researchers since the birth of the field \cite{reader_steps1969, turing_computing1950, harvey_fromanimals1991}. \citet{sigaud_definitionopenendedlearningproblems2024} and \citet{hughes_positionopenendedness2024} formalize open-endedness as the production of a sequence of novel and learnable artifacts, and the compositional and generalization properties of language make it a natural structure for generating such artifacts \cite{boyang_story2013, huang_visual2016, wang_chinese2016, santhanam_learning2020, zhu_visualize2023}. An agent able to autonomously select, generate, or discover goals is called autotelic \cite{colas_languageandculture2022}, a property \citet{sigaud_definitionopenendedlearningproblems2024} identifies as central to open-ended learning. 

Recent work uses LLMs to generate new goals by relabelling trajectories \cite{zhang_bootstrap2023}, or to create goals together with their reward functions \cite{wang_voyageropenendedembodiedagent2023, faldor_omniepic2025, pourcel_autotelic2024}. The goal spaces produced by such systems naturally exhibit the structure our method exploits (Section~\ref{methods}): a partial precedence relation between goals where some goals can be viewed as a composition of others. It appears explicitly in AutotelicLLM \cite{pourcel_autotelic2024}, whose reward functions are built around named subgoal sequences, and emergent in Voyager's tech-tree-like skill chains \cite{wang_voyageropenendedembodiedagent2023}. This relation is grounded in a small set of atomic primitives, and coexisting with infeasible or unlearnable goals that learnability criteria routinely filter out \cite{faldor_omniepic2025, pourcel_autotelic2024}. 

We assume such a mechanism has already produced goals and their reward functions, and focus on the complementary side of the open-ended loop: keeping goals learnable as their complexity grows. Exploring a vast goal space efficiently has been addressed by \textsc{MAGELLAN} \cite{gaven_magellan2025}, which selects goals suited to the agent's capabilities. Less addressed is the cost of acquiring each new goal in isolation, which grows rapidly with compositional depth \cite{adaptiveagentteam_humantimescaleadaptationopenendedtask2023, valmeekam_planningabilitieslargelanguage2023, poesia_learning2024}: without amortization of past learning, parts of the goal space remain out of reach within any practical budget. \herakles targets this side of the loop, enabling skill reuse and compositional generalization across an evolving goal space.

\section{The Herakles Method}
\label{methods}

The \herakles mechanism is based on a hierarchical agent composed of a high-level policy that directs a low-level policy, as depicted in Figure~\ref{fig:inference}. The latter aims to gradually internalize the behavior of the hierarchical agent. In this way, the hierarchical agent learns to solve a goal $g$ by relying on the low-level policy to achieve a set of sub-goals required for $g$. Once $g$ is considered mastered by the low-level policy, the high-level policy gains access to a new option—namely, invoking the low-level policy to execute $g$. This facilitates the achievement of more complex goals by reducing the corresponding signal-to-noise ratio during training \cite{park2023hiql}. Ultimately, this yields a global policy where the high-level component acts as a planner, focused on understanding situations and reasoning about composing skills to achieve goals, while the low-level component serves as a skill executor.

\begin{wrapfigure}{r}{0.45\textwidth}
\vspace{-1.5cm}
\captionof{algorithm}{\herakles (Sketched)}
%\caption{Compressed version of the \herakles algorithm a more comprehensive version is given in Appendix~\ref{app:herakles_algorithm}.}
\label{alg:herakles_compressed_version}
\begin{algorithmic}
\State $\text{Initialise empty buffers } \mathcal{B_{\text{HL}}}\text{, } \mathcal{B^{\text{h}}_{\text{LL}}}\text{, } \mathcal{B^{\text{sg}}_{\text{LL}}} $;
\State Set policies $\pihl$ and $\pill$; 
\State $\text{Competency estimator } C_{\theta}$;
\For{NbIterations}
\State $o \gets env.reset()$; Sample $g \sim P_G$; 
\State Set $\tau_g \leftarrow \emptyset $; $t \leftarrow 0$; $k \leftarrow 0$; $t_k \leftarrow 0$; %\emptyset$; % sample $g$
%\State $done_g, r^g \gets env.verify(g)$
\While{$(k < N_{HL})$ $\wedge$  $!done_g$}
\State Construct $\tilde{G}$ using $C_{\theta}$ ( Section~\ref{skill_space_generation})
\State Sample $\tilde{g} \sim \pihl^{\tilde{G}(o)}(\tilde{g}| \phi(o,g))$; 
%\State $done_{g}, r^{g} \gets env.verify(g)$ 
%\State $done_{\tilde{g}}, r^{\tilde{g}} \gets env.verify(\tilde{g})$
\State $t_k \leftarrow t$;  $k \leftarrow k +1$; $h \leftarrow 1$; 
\While{($t - t_k < N_{LL}$) $\wedge$  $!done_{\tilde{g}}$} 
\State $ a \gets \pill(\tilde{g},  o)$; 
%\State $env$ verify success of g and $\tilde{g}$
\State $o_{next}, r^{g}, r^{\tilde{g}} \gets env.step(a)$
%\State $done_{g}, r^{g} \gets env.verify(g)$ 
%\State $done_{\tilde{g}}, r^{\tilde{g}} \gets env.verify(\tilde{g})$
%\State $\mathcal{B^{\text{sg}}_{\text{LL}}} \gets (\tilde{g}, obs, a, r^{\tilde{g}})$ 
%\State $\mathcal{B^{\text{h}}_{\text{LL}}} \gets (g, obs, a, r^{g})$
%\State Policy compilation Section~\ref{rl_training}
%\If{$done_{\tilde{g}}$ or $ done_{g}$}
%\State $\mathcal{B_{\text{HL}}} \gets (g, %(obs,\tilde{G}), a, r^{g})$
%\EndIf
\State $\tau_g \leftarrow \tau_g \oplus (o, h, \tilde{g}, a, r^{g}, r^{\tilde{g}})  $; 
\State $o \gets o_{next}$; $t \leftarrow t + 1$; $h \leftarrow 0$;
\EndWhile
\EndWhile
\State $\tau_g \leftarrow \tau_g \oplus o$;
\State Insert $(\tau_g,g)$ in buffers (Section~\ref{rl_training});  
\State Update $\pi^{LL}$ on $\mathcal{B^{\text{h}}_{\text{LL}}} \cup \mathcal{B^{\text{sg}}_{\text{LL}}}$;
\State Update $C_{\theta_k}$ on $\mathcal{B^{\text{sg}}_{\text{LL}}}$;
\State Update $\pi^{HL}$ on $\mathcal{B_{\text{HL}}} $;
\EndFor
\end{algorithmic}
%\vspace{-1.25cm}
\end{wrapfigure}

\subsection{Problem statement}
\label{problem_statement}

Let $\mathcal{M}=(S, A, \mathcal{T}, G, r, \Omega, O, \gamma)$ be a goal-augmented Partially Observable Markov Decision Process, with $S$ the state space, $A$ the primitive action space, $\mathcal{T}$ the transition function, $G$ a pre-generated goal space with an associated sparse reward function $r: S \times A \times G \rightarrow \{0; 1\}$, $\Omega$ the observation space, $O: S \rightarrow \Omega$ the observation function that maps states to observations, and $\gamma \in [0, 1)$ the discount factor.  Given an initial state distribution $P_{s_0}$ over $\mathcal{S}$ and a goal distribution $P_\mathcal{G}$ over $\mathcal{G}$, we seek a goal-conditioned policy $\pi: \Omega \times \mathcal{G} \rightarrow \Delta(\mathcal{A})$ maximizing the expected discounted return $\pi^* \;=\; \arg\max_{\pi} \, \mathbb{E}_{s_0 \sim P_{s_0},\, g \sim P_\mathcal{G}} \mathbb{E}_{\tau \sim \pi(\cdot \mid s_0, g)}, \bigl[ R_g(\tau) \bigr]$, where $\pi(\tau \mid s_0, g)$ denotes the probability of trajectory $\tau$ under $\pi$ conditioned on goal $g$ and starting state $s_0$, and the goal-conditioned return is defined as $R_g(\tau) \,=\, \sum_{t=0}^{|\tau|} \gamma^{t}\, r_t^{g}$,  with $r_t^{g} = r(s_t, a_t, g)$. % and $(s_t, a_t)$ is the state–action pair at step $t$ of $\tau$.
%Depending on the setting, $P_G$ can be defined as a static distribution over a given goal space, or can evolve along with competencies of an open-ended exploration strategy relying on the agent.   
%Depending on the setting, $P_\mathcal{G}$ may either be static or evolve over time to reflect the distribution of goals that policies encounter during open-ended exploration. 
Depending on the setting, $P_\mathcal{G}$ may either be static or evolve over time according to an exogenous process reflecting the distribution of goals encountered during open-ended exploration. 
In order to guide the agent and extract semantic relationships between goals,  we also consider that the agent  has access to a prompting function $\phi:\Omega \times G \rightarrow {\cal V}^{L_p}$ that transforms any pair (observation, goal) into a textual prompt of $L_p$  tokens from a given vocabulary ${\cal V}$.

Let the goal space $G$ be endowed with a precedence relation  $\prec$, such that for some pairs  of goals $(g',g) \in G^2$, $g' \prec g$ indicates that achieving $g'$ is a prerequisite for the completion of $g$. 
We assume that this relation defines a hidden  hierarchical structure over \(G\), whereby complex goals can be decomposed into prerequisite subgoals. Accordingly, for any pair \((g', g)\) satisfying \(g' \prec g\), any trajectory \(\tau_g = \{s_0, a_0, s_1, a_1, \dots, s_n\}\) that achieves \(g\) can be decomposed into a prefix trajectory \(\tau_{g'}\) that achieves \(g'\) at some step $k \in (0,n)$, followed by a suffix  \(\tau_{g' \rightarrow g} = \{s_k, a_k, \dots, s_n\}\) leading from \(g'\) to \(g\). To anchor $\prec$, we define a minimal set of atomic goals by associating, with each primitive action $a \in \mathcal{A}$, an action-matching goal $g_a \in \mathcal{G}$ such that $R(s, a, g_a) = 1$ for all $s \in \mathcal{S}$. This is a mild requirement: it amounts to naming each primitive action as a goal, and guarantees that the minimal elements of $\prec$ are reachable in a single step. %Any goal connected to this base through $\prec$ then admits a finite decomposition into primitive-level subgoals. 
In the following, we exploit this latent structure by assuming a correspondence between linguistic compositionality, goals expressed as natural-language prompts through $\phi$, and skill compositionality (similarly to \cite{ahn_icanisay2022, ahuja_hierarchicalreinforcementlearningnatural2023}), allowing the agent to discover and reuse prerequisite chains wherever they exist in $\mathcal{G}$.

\begin{figure}
\vspace{-\baselineskip}
    \centering
    \includegraphics[width=0.7\linewidth]{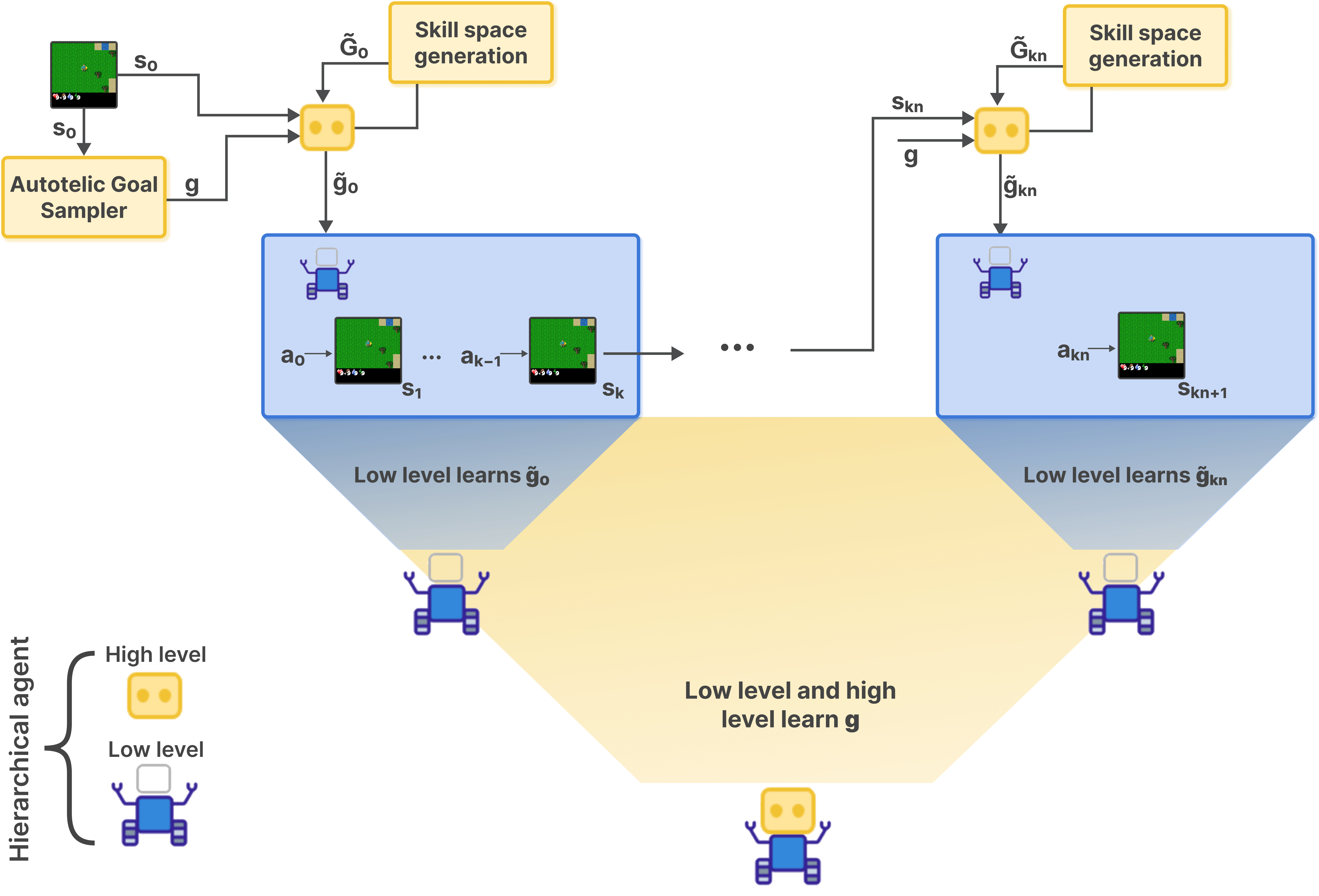}
    \caption{{\bf Skill learning and compilation in {\bf HERAKLES}}: Given a goal $g$ and a state $s_0$, $\pihl$ generates its set of options $\tilde{G}$ and samples one option $\tilde{g}_0$, where an option corresponds to a goal from $G$ that is considered as mastered by the LL policy. This option is executed by $\pill$ using primitive actions in the environment, leading to state $s_k$. Given $s_k$ and $g$, $\pihl$ then generates a new set of options, samples a new option $g_k$ and passes it to $\pill$. This process repeats until $g$ has been achieved or the limited number of high-level steps has been reached. $\pihl$ is trained on the high-level trajectory consisting in the selection of options. $\pill$ is trained on all the low-level trajectories (one for each option) and on the \textbf{compiled trajectory} that is the concatenation of all the low-level trajectories to reach the goal $g$ passed to the hierarchical agent. Doing so, $\pill$ progressively internalizes the sequence of primitive actions required to directly solve $g$.}
\label{fig:learning_and_compilation}
\vspace{-\baselineskip}
\end{figure}

\subsection{Hierarchical Compilation}

To exploit the compositional structure of $G$, we consider a hierarchical agent $\pi = (\pihl, \pill)$ operating at two levels of temporal abstraction. The low-level policy $\pill : \Omega \times G \rightarrow \Delta(A)$ is a goal-conditioned controller mapping observations and goals to primitive actions. The high-level policy $\pihl : \phi(\Omega, G) \rightarrow \Delta(G)$, implemented as a language model, selects subgoals based on textual prompts. At execution time, $\pihl$ proposes subgoals that are executed by $\pill$ until termination, inducing a Semi-Markov Decision Process (SMDP). A subgoal terminates when it is achieved or after $N_{\mathrm{LL}}$ steps, while the high-level policy selects at most $N_{\mathrm{HL}}$ subgoals.

A trajectory for a goal $g$ is defined as:
$
\tau_g = (s_0, h_0, g_0, a_0, r_0^g, r_0^{g_0}, \dots, s_T),
$ where $s_t$ is the state, $h_t \in \{0,1\}$ indicates high-level decision steps ($h_t=1$ triggers a query of $\pihl$), $g_t \in G$ is the subgoal executed by the low-level policy, and $a_t$ is the primitive action. All trajectories respect $h_0=1$, as we impose the agent starting %assume the trajectory starts 
with a high-level call. Each step contains two rewards: $r_t^g$ 
 is the global (high-level) reward, while $r_t^{g_t}$  is the reward associated with the current subgoal $g_t$ (used for off-policy training of $\pill$). % reward densification during training). 
 Let $I(\tau_g)=\{t_k\}_{k=0}^{K-1}$ be the ordered set of steps where $h_t=1$. The projected trajectory over high-level decision steps is thus: $
P_g(\tau_g) = \big((s_{t_k}, g_{t_k}, R_k^g)\big)_{k=0}^{K-1} \oplus (s_T),
$ 
with $\oplus$ the concatenation operator and 
$ 
R_k^g = \sum_{t=t_k}^{t_{k+1}-1} \gamma^{\,t - t_k} \, r_t^g.
$ 
Let us define $\gamma_{\mathrm{HL}} = \gamma^{c_{\mathrm{HL}}}$ as the high-level discount factor, with $c_{\mathrm{HL}}$ a cost for high-level decisions. Given a low level policy $\pill$, the high-level (SMDP) objective is:
\begin{equation}
\pihl^* = \arg\max_{\pihl}
\mathbb{E}_{(s_0,g)} \mathbb{E}_{\tau_g \sim (\pihl, \pill)}
\left[ \sum_k \gamma_{\mathrm{HL}}^k R_k^g \right].
\label{hl_obj}
\end{equation}

\paragraph{Bilevel structure of the optimal policy}

Unlike standard hierarchical RL, where subgoals are transient and defined by the high-level policy, our setting considers a shared goal space $G$ in which both task goals and intermediate subgoals reside, enabling reusable skill learning and explicit compositionality (as we assume a precedence relation between goals $\prec$ exists). Concretely, $\pill$ stands as a universal goal-conditioned policy expected to directly solve a large subset of goals from states reached by the hierarchical policy.  Consequently, this objective does not define a single-level policy search, but induces a bilevel coupling between $\pihl$ and $\pill$ through the state-subgoal distribution induced by $\pi=(\pihl, \pill)$, defined as $d_h^\pi(s,g') = \mathbb{E}_{g \sim P_G, s_0 \sim P_{s_0}}
\left[
P^\pi(s_t = s \mid h_t=1, s_0, g)\;
\pihl(g' \mid \phi(s, g))
\right]$.  
In particular, given $\pihl$, the optimal $\pill$ is given as:
\begin{equation}
\pill^*
=
\arg\max_{\pill}
\mathbb{E}_{(s,g) \sim d_h^{(\pihl, \pill)}}
\mathbb{E}_{\tau \sim \pill(\cdot|s,g)}
\left[
\sum_t \gamma^t r_t^g
\right].
\label{ll_obj}
\end{equation}

\paragraph{Coupled optimality}

%The overall optimal policy is thus characterized by a coupled fixed-point structure:
We can interpret the joint optimum as a fixed point of coupled best-response operators:
%\vspace{-0.5cm}
\[
\pi^* = (\pihl^*, \pill^*)
\quad \text{such that} \quad
\begin{cases}
\pihl^* \text{ maximizes \eqref{hl_obj} given } \pill^* \\
\pill^* \text{ maximizes \eqref{ll_obj} given } \pihl^*.
\end{cases}
\]

%This coupling reflects the fact that $\pihl$ shapes the distribution of states and goals seen by $\pill$, while the effectiveness of low-level execution determines the optimality of high-level decomposition.
%Finally, $\pihl$ both guides exploration through structured subgoal decomposition and composes low-level skills when necessary. The resulting policy is therefore hybrid: in the limit of full compilation execution  reduces to a single high-level decision for simple goals, and falls back to hierarchical sequencing for more complex ones. 
%The parameter $c_{\mathrm{HL}}$ controls the trade-off between these two regimes. By penalizing high-level decisions, it encourages temporally extended behaviors in $\pill$, promoting the compilation of skills, while still allowing hierarchical reasoning when required. Importantly, $\pihl$ remains part of the final policy, acting as a residual composer for goals that cannot be directly solved by $\pill$.

This Stackelberg-like coupling \cite{fiez_convergence2019, shen_principled2024, chakraborty_parl2024} reflects that $\pihl$ shapes the distribution of states  and subgoals seen by $\pill$, while low-level execution determines the optimality of high-level decomposition. Beyond this fixed-point view, $\pihl$ guides exploration through subgoal decomposition and composes low-level skills when needed, yielding a hybrid policy: in the limit of full compilation, simple goals reduce to a single high-level decision, while complex ones fall back to hierarchical sequencing. The parameter $c_{\mathrm{HL}}$ controls this trade-off---penalizing high-level decisions encourages temporally extended behaviors in $\pill$ and promotes skill compilation, while still allowing hierarchical reasoning. Importantly, $\pihl$ remains part of the final policy, acting as a residual composer for goals that $\pill$ cannot directly solve.

\subsection{RL training of the HL and LL policies}
\label{rl_training}

Given the coupled optimality defined in previous  section, %\herakles, 
$\pihl$ and $\pill$ are learned simultaneously (see Figure~\ref{fig:learning_and_compilation}), following the sketched  Algorithm~\ref{alg:herakles_compressed_version}. %A detailed version of the full
The full training procedure is given in Appendix~\ref{app:herakles_algorithm}. 

In the following, let $\mathcal{D}$ denote a dataset of recent trajectories paired with their task goals. From $\mathcal{D}$, we construct three replay buffers corresponding to different levels of abstraction. The high-level buffer contains projected trajectories at the high-level decision scale:  
$
\mathcal{B}_{HL} := \{ (P_g(\tau_g), g) \mid (\tau_g, g) \in \mathcal{D} \}
$. 
The low-level buffer of hierarchical rollouts contains full (flattened) trajectories labelled by the task goal: 
$
\mathcal{B}_{LL}^{h} := \{ ((s_t, a_t, r_t^g)_{t=0}^{T-1} \oplus s_T, g)
\mid (\tau_g, g) \in \mathcal{D} \}$. 
Finally, the subgoal-conditioned buffer contains segments between high-level decisions: 
$
\mathcal{B}_{LL}^{sg} := \{ ((s_t, a_t, r_t^{g_{t_k}})_{t=t_k}^{t_{k+1}-1} \oplus s_{t_{k+1}}, g_{t_k})  \mid (\tau_g, g) \in \mathcal{D},\ t_k \in I(\tau_g) \}
$.

\paragraph{High-level policy training}

The high-level policy $\pihl$ is implemented as a pre-trained language model equipped with a LoRA adapter~\cite{Hu_LoRA2021}. This adapter is fine-tuned using on-policy RL over trajectories sampled from $\mathcal{B}_{HL}$.

\paragraph{Low-level policy training}

The low-level policy $\pill$ is a lightweight neural network operating on raw observations. It is trained off-policy using both $\mathcal{B}_{LL}^{h}$ and $\mathcal{B}_{LL}^{sg}$, which serve complementary roles.

\smallskip
\noindent
\textbf{Type 1 — Hierarchical trajectories.}
Samples from $\mathcal{B}_{LL}^{h}$ correspond to full rollouts generated by the hierarchical policy. These trajectories provide supervision at the task level, allowing $\pill$ to \emph{compile} successful hierarchical behaviors into direct goal-conditioned policies using only primitive actions.

\smallskip
\noindent
\textbf{Type 2 — Subgoal trajectories.}
Samples from $\mathcal{B}_{LL}^{sg}$ correspond to segments collected during the execution of subgoals. These trajectories provide denser and more localized feedback, enabling $\pill$ to reinforce previously acquired skills and maintain performance across the goal space, while training it on states induced by high-level decisions, i.e., samples from the distribution $d_h^\pi$ induced by the current policy.

\smallskip
Together, both training signals allow $\pill$ to acquire new skills through behavioral compilation and to stabilize existing ones through repeated subgoal execution. 
As the hierarchical agent successfully solves a goal $g$, the corresponding behaviors are progressively compiled into $\pill$. Consequently, $\pill$ becomes able to directly reach $g$ from relevant states, enabling $\pihl$ to reuse it as a primitive for more complex goals. 
As training progresses, a growing fraction of execution is delegated to $\pill$, reducing reliance on high-level decisions. This improves efficiency and facilitates exploration for $\pihl$, which operates over temporally extended and more predictable transitions.

\subsection{Adaptive Subgoal Filtering for the High-Level Policy}
\label{skill_space_generation}

The general principle of our approach is to  progressively acquire new skills that can be compiled into $\pill$. Once reliably executable, these skills form higher-level actions that $\pihl$ can invoke to more efficiently achieve its goals.

Rather than allowing $\pihl$ to consider all subgoals in $G$, which is inefficient when only a subset is achievable by $\pill$, we introduce a pre-filtering step, as depicted in Figure~\ref{fig:inference} which illustrates the hierarchical decision process. This filtering step restricts the set of admissible subgoals to $\tilde{G}(o) \subseteq G$, containing only goals that the low-level policy is expected to reliably execute from the current state, given its observation $o$. Given this restricted set, $\pihl$ is implemented as an LLM agent that selects subgoals from $\tilde{G}(o)$ as options to trigger. Following \cite{carta_grounding2023}, the probability of sampling a subgoal $\tilde{g} \in \tilde{G}(o)$, with textual description $\phi(\tilde{g})$, given a goal $g$ and observation $o$, is defined as:
$
\pihl(\tilde{g} \mid \phi(o,g)) \propto P_{LLM}(\phi(\tilde{g}) \mid \phi(o,g)),
$ 
where $P_{LLM}(\cdot \mid \phi(o,g))$ denotes the probability induced by the LLM under constrained decoding, restricted to sequences corresponding to valid goal descriptions in $\tilde{G}(o)$. For a given goal $g$ and observation $o$, we denote as 
$\pihl^{\tilde{G}(o)}(.|\phi(o,g))$ our HL policy restricted to subgoals from $\tilde{G}(o)$.   

\paragraph{Constructing $\tilde{G}_k$(.)} 
A crucial aspect of our method is how to define, at any state reached by the high-level controller, the set of callable options compatible with the \emph{current} capabilities of $\pill$. Constructing $\tilde{G}(o)$ requires balancing two competing objectives: (i) avoiding exposure of $\pihl$ to options not yet mastered by $\pill$, which would destabilize learning, and (ii) introducing new options to $\pihl$ as early as possible so that it can rapidly adapt to them. Let $\tilde{G}_k(o)$ denote the set $\tilde{G}(o)$ at iteration $k$ of the learning process. We can bootstrap $\tilde{G}_0(\cdot)$ as the set containing only the primitive actions, which by definition $\pill$ always masters, by using the fact that $G$ contains at least one goal $g_a$ for each primitive action $a \in A$.

The set of available actions for $\pihl$ is built at any step $k$ of the learning process based on a competency estimator $C_{\theta_k}$, that we learn from a buffer of low-level trajectories  $\mathcal{B^{\text{sg}}_{\text{LL}}} $ 
(see section \ref{rl_training}) by minimizing the binary cross-entropy loss: $$\mathcal{L}(\theta_k) = \mathbb{E}_{(\tau, g) \in 
\mathcal{B^{\text{sg}}_{\text{LL}}}} 
\mathbb{E}_{o \in \tau} \left[ BCE(I(R_g(\tau)>0)
, C_{\theta_k}(o, g)) \right],$$ with 
$\mathcal{B^{\text{sg}}_{\text{LL}}}$  
containing recent pairs $(\tau,g)$, with $\tau$  any sub-trajectory sampled from  the distribution $\pill(.|g)$ of trajectories following the LL policy conditioned on goal $g$ (i.e., starting from an HL call to $g$). $I(R_g(\tau)>0)$ is the indicator function that returns 1 if goal $g$ is achieved in $\tau$, 0 otherwise. 
Similarly to 
\cite{gaven_magellan2025}, $C_{\theta_k}(o,g)$ is defined as an MLP on top of the representation produced by an LLM for the last token of the prompt $\phi(o,g)$. $\theta_k$ stands as parameters of this MLP and those of a specific LORA \cite{Hu_LoRA2021} adapter of a base LLM. 

Given an observation $o_t$, $C_{\theta_k}(o_t, g)$ predicts the probability that $\pill$ successfully achieves $g$ from $o_t$. At any decision step $t$ where $h_t = 1$, each goal $g \in G$ is independently included in $\tilde{G}_k(o_t)$ via a Bernoulli draw with probability: $p_g = \max\big(C_{\theta_k}(o_t, g), \epsilon\big),$where $\epsilon$ is an exploration hyper-parameter. This construction directly addresses both objectives: $C_{\theta_k}(o_t, g)$ keeps the inclusion probability low for goals $\pill$ has not yet mastered from $o_t$, satisfying~(i); meanwhile, the floor $\epsilon$ guaranties that every goal retains a minimal selection probability, ensuring that new or under-practiced goals are still occasionally proposed to $\pihl$, satisfying~(ii). The exploration parameter $\epsilon$ avoids a self-reinforcing loop in which unmastered goals receive low scores, are never practiced, and consequently prevent both $\pill$ and $C_{\theta_k}$ from improving on them.

\section{Experiments}
\label{experiments}
\subsection{Experimental setup}
\begin{figure}[h]
\vspace{-\baselineskip}
    \centering
    
    \begin{minipage}{0.48\linewidth}
        \centering
        \includegraphics[width=\linewidth]{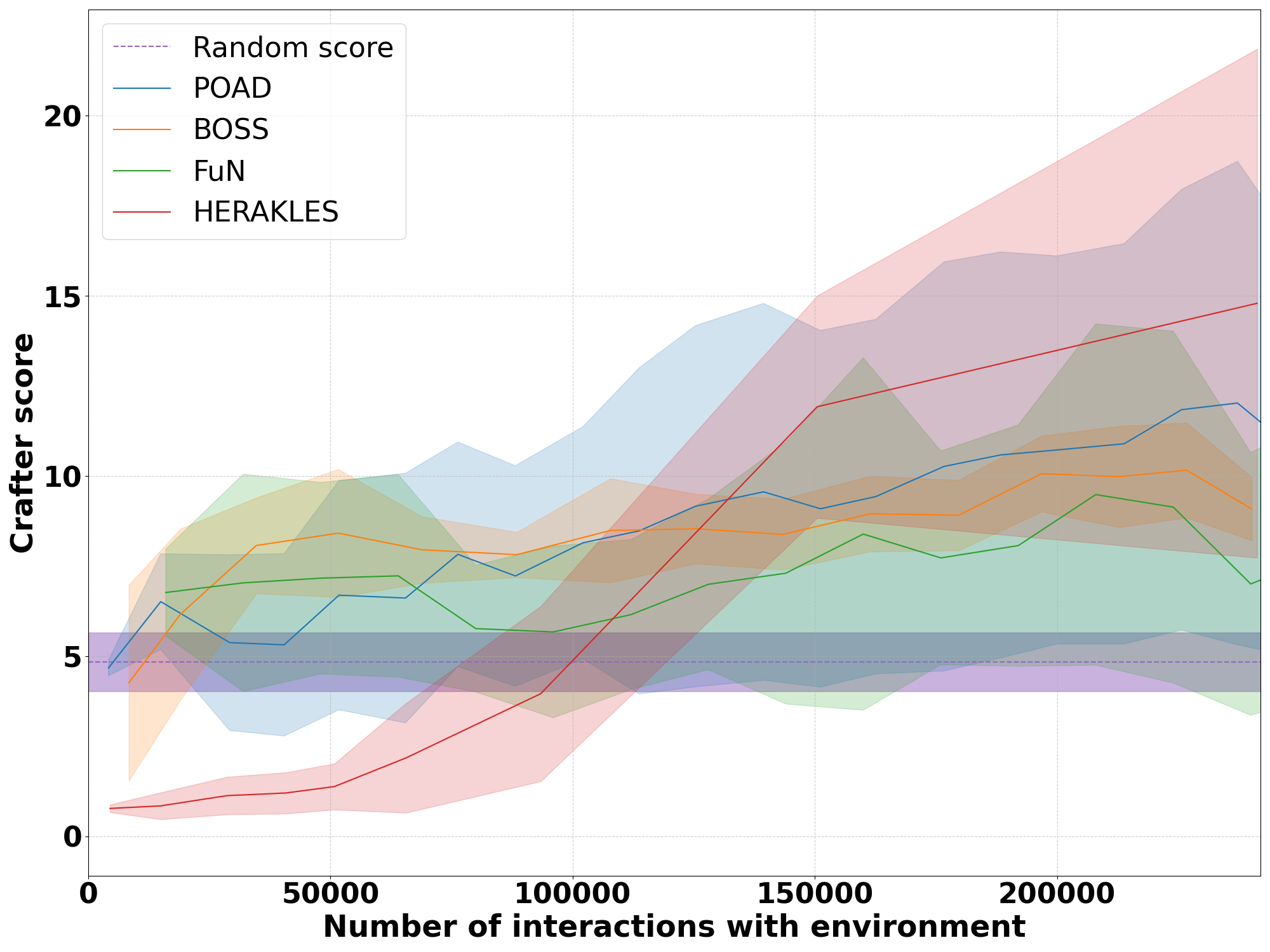}
        \caption{Crafter score as a function of the number of steps in the environment. Shaded area represent standard deviation using $5$ seeds.}
        \label{fig:evolution_crafter_score_interaction_steps}
    \end{minipage}
    \hfill
    \begin{minipage}{0.48\linewidth}
        \centering
        \includegraphics[width=\linewidth]{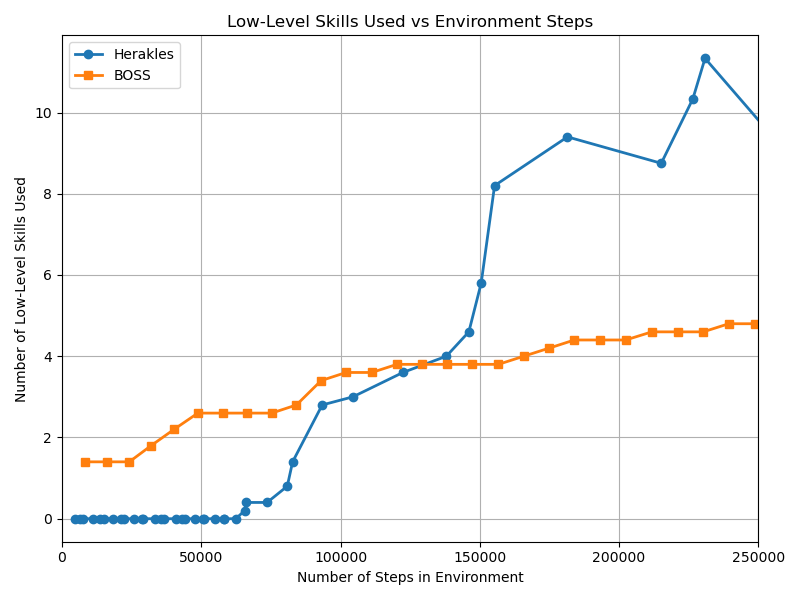}
        \caption{Evolution of the average number (over $5$ seed) of low level skills that $\pihl$ can call depending on the number of steps.}
        \label{fig:nbr_skill_used}
    \end{minipage}
\vspace{-\baselineskip}
\end{figure}

\paragraph{Environment description}
We evaluate our method in Crafter \cite{hafner_benchmarking2022} using the version proposed by \cite{du_guiding2023} (more details in Appendix~\ref{app:modifications}). It is a procedurally generated and partially observable Minecraft-like 2D environment that enables collecting and creating a set of artifacts organized along an achievement tree of $17$ goals (see Appendix~\ref{app:achievement_tree}). Most of the goals require the agent to reuse discovered or generated artifacts from previous goals. For instance,  to "\textit{make a wood pickaxe}", the agent needs to have built a crafting table. While the Crafter goal space is finite, it captures a locally open-ended setting by presenting an expanding set of learnable goals through environment interactions. This property enables us to study the scaffolding dynamics of hierarchical agents—specifically, how the high-level policy mobilizes and recombines an increasingly rich repertoire of low-level skills to achieve progressively complex goals. 
Consequently, the Crafter goal space is an excellent candidate for testing our method’s ability to identify and exploit the underlying goal structures within a goal space. Such structures are found in the goal spaces generated by open-ended agents\cite{hughes_positionopenendedness2024}\cite{wang_voyageropenendedembodiedagent2023}.%Consequently, Crafter's goal space is an excellent candidate for simulating goal spaces produced by open-ended agents \cite{hughes_positionopenendedness2024}. 
For our experiments, we equip the environment with a prompting function $\phi$, that generates textual description of the scene. Thus, an observation $o \in \Omega$ is a tuple made from visual and textual observations $o=(o_v, o_{text})$ (see Appendix~\ref{app:observations} for more details). $\pihl$ receives the linguistic description $o_{text}$ and $\pill$ uses the visual observation $o_v$.  

\paragraph{Goal Sampler}
%In Crafter, the agent is encouraged to explore and master as many goals as possible. The difficulty of a goal $g \in G$ is not fixed, but conditioned on the current state of the environment $s_k$. For example, the goal \textit{'place table'} becomes substantially easier if the agent already possesses wood in its inventory. To enable the agent to select a goal $g$ conditionally to a state $s_T$ at a time step $T$ that maximizes the hierarchical policy competence across the large goal space $G$ of Crafter, we draw inspiration from methods based on Learning Progress (LP). These approaches prioritize goals that are expected to yield the greatest future improvement through practice~\citep{oudeyer_intrinsic_2007}. Specifically, we implement the MAGELLAN algorithm~\cite{gaven_magellan2025}, which equips LLM agents with the ability to predict and generalise their current competence and learning progress online over the goal space by leveraging semantic relationships between goals. Implementation details are provided in Appendix~\ref{app:magellan_implementation}.

In Crafter, the agent is encouraged to master as many goals as possible. Goal difficulty 
is not fixed but conditioned on the current environment state $s_k$. For instance, 
\textit{`place table'} is substantially easier if the agent already carries wood. 
To select goals that maximize hierarchical policy competence across the large goal 
space $G$, we draw inspiration from Learning Progress (LP) methods, which prioritize 
goals expected to yield the greatest future improvement~\citep{oudeyer_intrinsic_2007}. 
Specifically, for all approaches compared in the experiments, we use the MAGELLAN algorithm~\cite{gaven_magellan2025} to define $P_G$.  This goal sampling strategy, based on LLM models to predict and generalize competences across the goal space, simulates the evolution of the goal distribution that open-ended agents’ policies may encounter within an open-ended exploration setting. %This goal sampling strategy, based on LLM models to predict and generalize competences across the goal space, %and learning progress online distribution 
%simulates the evolution of the goal distribution that open-ended agents % space  which enables 
%LLM agents to predict and generalize competence and learning progress online by 
%leveraging semantic relationships between goals. 
Implementation details are provided in Appendix~\ref{app:magellan_implementation}.

\paragraph{Training HL and LL policies}
We model $\pihl$ using Mistral 7B and train it with the \textsc{poad} on-policy RL algorithm \cite{wen_reinforcing2024}. %At each step of the high-level we sample a skill inside the current skill space using constrained decoding. 
For $\pill$, we adapt the ResNet-based architecture given by \cite{moon_discovering2023} ($2M$ parameters). We train $\pill$ using the AWR off-policy algorithm \cite{peng_advantageweightedregression2019} and fill the buffer with both types of trajectories described in Section~\ref{rl_training}. We approximate the timescale separation required for convergence of the Stackelberg-like coupling by training the lightweight $\pill$ faster than the LLM-based $\pihl$. Appendices~\ref{app:high_level_policy} and \ref{app:low_level_policy} give more details on $\pihl$, $\pill$, and their training.

\paragraph{Baselines}

To measure the impact of HRL in \herakles, we compare with \textsc{poad}~\cite{wen_reinforcing2024}, 
which corresponds to the high-level policy of \herakles restricted to elementary actions $A$ 
(see Appendix~\ref{app:poad_training}). To evaluate the benefit of a language-structured skill space, 
we compare with \textsc{FuN}~\cite{vezhnevets_feudalnetworkshierarchicalreinforcement2017}, a feudal 
HRL method where the high-level policy generates subgoals in a latent embedding space. To match 
\herakles's automatic curriculum, we make \textsc{FuN}'s high-level policy goal-conditioned 
by appending a goal embedding (produced by the same LLM as \herakles) to the current 
observation (see the Appendix~\ref{app:FuN}). Finally, we compare against \textsc{BOSS}~\cite{zhang_bootstrap2023}, 
which also uses an LLM as a frozen high-level policy to compose skills hierarchically, but relies on a 
pre-trained skill library initialized from demonstrations and chains skills bottom-up via trajectory relabelling, in contrast to \herakles's continuous skill compilation. 
Training details and adaptations are provided in Appendix~\ref{app:boss_training}. In Appendix~\ref{app:complementary_results}, we present further experiments and more details about the effects of our methods.

\subsection{Sample efficiency}

We train our agents for $250{,}000$ interaction steps in Crafter and report the Crafter score~\citep{hafner_benchmarking2022}:
$S_c = \exp\bigl(\frac{1}{N} \sum^N_{i=1} \ln(1 + sr_i)\bigr)-1,$ where $sr_i \in [0; 100]$ is the success rate on goal $i$ and $N = 17$. The
geometric mean emphasizes rare and difficult achievements, yielding a
difficulty-aware metric without prior knowledge of goal complexity. Success
rates are estimated on $40$ held-out environments at regular intervals.

Figure~\ref{fig:evolution_crafter_score_interaction_steps} shows that baselines initially progress faster than \herakles\ but plateau after mastering easy goals (e.g.\ \textit{go to tree}, \textit{place table}), making only marginal progress on harder ones. In contrast, \herakles\ improves gradually as more goals are compiled into $\pill$ (Figure~\ref{fig:nbr_skill_used}), enabling the hierarchical policy to reach complex goals and steadily increase its score. \textsc{BOSS} quickly masters a few easy low-level skills but stagnates, failing to discover more complex ones through its exploration process: upon manual inspection, the most complex skill reached by one of its seeds is \textit{place table}, against \textit{make stone pickaxe} for \herakles. Further analyses are reported in Appendix~\ref{app:evolution_of_success_rate}.

\subsection{Generalization to new goals}

\begin{table}[htbp]
\vspace{-\baselineskip}
\caption{Success rates on unseen goals. Goals with a star (*) were seen during training. We give the average results and the standard deviation over $5$ seeds. More results are given in Appendix~\ref{app:generalization_results}.}

\begin{adjustbox}{width=\textwidth}
\begin{tabular}{llcccc}
\toprule
\textbf{Generalization} & \multirow{2}{*}{\textbf{Goal}} & \multirow{2}{*}{\textbf{HERAKLES}} & \multirow{2}{*}{\textbf{FuN}} & \multirow{2}{*}{\textbf{BOSS}} & \multirow{2}{*}{\textbf{POAD}} \\
\textbf{Type}& & & & & \\
\midrule
\multirow{4}{*}{Compositional} 
  & collect wood*       & $\bm{0.92 \pm 0.05}$ & $0.73 \pm 0.18$ & $0.88 \pm 0.07$ & $0.80 \pm 0.39$ \\
  & collect 2 woods     & $\bm{0.92 \pm 0.05}$ & $0.35 \pm 0.15$ & $0.72 \pm 0.07$ & $0.41 \pm 0.48$ \\
  %& collect 3 woods     & $\bm{0.93 \pm 0.06}$ & $0.17 \pm 0.05$ & & $0.21 \pm 0.42$ \\
  & collect 4 woods     & $\bm{0.91 \pm 0.06}$ & $0.08 \pm 0.07$ & $0.55 \pm 0.08$ & $0.07 \pm 0.27$ \\
\midrule
\multirow{4}{*}{Synonyms} 
  & collect wood*       & $\bm{0.92 \pm 0.05}$ & $0.73 \pm 0.18$ & $0.88 \pm 0.07$ & $0.80 \pm 0.39$ \\
  & acquire wood        & $\bm{0.92 \pm 0.05}$ & $0.80 \pm 0.19$ & $0.87 \pm 0.02$ & $0.79 \pm 0.02$ \\
  %& place table*        & $\bm{0.73 \pm 0.12}$ & $0.30 \pm 0.15$ & $0.78 \pm 0.08$ & $0.05 \pm 0.23$ \\
  %& install table       & $\bm{0.70 \pm 0.11}$ & $0.32 \pm 0.15$ & & $0.10 \pm 0.30$ \\
  & make wood pickaxe*  & $\bm{0.47 \pm 0.27}$ & $0.11 \pm 0.06$ & $0.37 \pm 0.08$ & $0.75 \pm 0.07$ \\
  & create wood pickaxe & $\bm{0.44 \pm 0.39}$ & $0.16 \pm 0.20$ & $0.39 \pm 0.07$ & $0.71 \pm 0.11$ \\
\midrule
\multirow{2}{*}{\shortstack[l]{Similar trajectory\\ in skill space}} 
  & make wood pickaxe*  & $\bm{0.47 \pm 0.27}$ & $0.11 \pm 0.06$ & $0.37 \pm 0.08$ & $0.75 \pm 0.07$ \\
  & make wood sword & $\bm{0.40 \pm 0.37}$ & $0.00 \pm 0.00$ & $0.33 \pm 0.05$ & $0.00 \pm 0.05$ \\
\bottomrule
\end{tabular}
\end{adjustbox}
\label{table:generalization}
\vspace{-\baselineskip}
\end{table}

As \herakles uses an LLM as a high-level policy, we examine its ability to generalize to three types of new goals:
\par\smallskip $\bullet$ \textbf{Compositional generalization}: goals repeating a known goal $k$ times (\emph{e.g.}, \textit{`collect $k$ woods'}).
\par\smallskip $\bullet$ \textbf{Synonymous goals}: goals in which verbs are replaced 
by synonyms.
\par\smallskip $\bullet$ \textbf{Trajectory-similar goals}: we evaluate on the unseen goal \textit{`make wood sword'} which shares the same prerequisite as the training goal \textit{`make wood pickaxe'}, both require crafting a table and holding wood, differing only in the final elementary action (\textit{craft} \texttt{wood sword} vs.\ \textit{craft} \texttt{wood pickaxe}).

Table~\ref{table:generalization} reports success rates on unseen goals after $250{,}000$ training steps; pretrained goals are marked with a star (*). All language-based methods generalize to varying degrees, but \herakles suffers the least: it shows a negligible drop from \textit{‘collect wood’} to \textit{‘collect $4$ woods’}, unlike the baselines. The ‘Synonyms’ row shows all methods handle synonymous verbs, yet baselines fail to reuse skills for non-synonymous goals sharing a similar trajectory. In contrast, \herakles leverages the linguistic structure of its skill space, retaining strong performance ($0.47 \rightarrow 0.40$). Additional generalisation results are in Appendix~\ref{app:generalization_results}.

\section{Conclusion}
\label{sec:conclusion}

We introduced \herakles, a hierarchical framework for open-ended autotelic agents in which an LLM-based high-level policy and a lightweight low-level policy are trained concurrently. As the agent masters increasingly complex goals, the corresponding behaviours are compiled into the low-level policy as reusable skills for  %alearned
the high-level policy. This dynamic skill expansion enables effective operation in evolving goal spaces without expert-defined skills or pre-training. In the Crafter environment, \herakles scales with goal complexity, improves sample efficiency over prior HRL and LLM-only approaches, and generalizes to compositional, synonymous, and trajectory-similar goals. Several limitations remain. The high-level policy operates on textual goals only; extending it to visual goals via Vision-Language Models~\citep{liu2024survey} would broaden its applicability. The low-level policy's periodic retraining on previously learned skills also incurs computational overhead we would like to reduce. Finally, our setting assumes an externally provided goal space; integrating \herakles with goal-generation~\citep{faldor_omniepic2025, pourcel_autotelic2024} and skill-discovery methods is a promising direction toward fully autonomous open-ended agents.

\begin{ack}
This work benefitted from access to the HPC resources of IDRIS under the allocation A0171011996 made by GENCI. It was also co-funded by AI Chair ANR DeepCuriosity ANR-19-CHIA-0004. C
\end{ack}

\small
\bibliography{biblio}

\newpage

\appendix

\section*{Appendices}

This supplementary material provides additional results, discussion, and implementation details.

\begin{itemize}
    \item Section~\ref{app:env_detail} details our environment. 
    \begin{itemize}
        \item Section~\ref{app:achievement_tree} provides our environment's achievement tree with the list of all possible goals and their requirements.
        \item Section~\ref{app:modifications} details the modifications we performed to Crafter.
        \item Section~\ref{app:observations} details the observation space used by our agents.
    \end{itemize}
    \item Section~\ref{app:complementary_results} provides complementary results. 
    \begin{itemize}
        \item Section~\ref{app:evolution_of_success_rate} shows the per-task success rate for each method across training.
        \item Section~\ref{app:generalization_results} analyzes the generalization abilities of each method. We study both generalization to synonyms and n-compositionality.
        \item Section~\ref{app:evolution_hl_sampling_strategy} provides insights on \herakles' high-level sampling strategy. We notably study the evolution fo sampling strategy for multiple goals.
    \end{itemize}
    \item Section~\ref{app:hierarchical_agent} details \herakles' interactions between the high-level and low-level policy.
    \begin{itemize}
        \item Section~\ref{app:herakles_algorithm} provides \herakles' full algorithm.
        \item Section~\ref{app:herakles_implementation} gives details about our implementation of \herakles. We also provide the link to our code.
    \end{itemize}
    \item Section~\ref{app:high_level_policy} details the implementation of our high-level policy.
    \begin{itemize}
        \item Section~\ref{app:high_level_policy_architecture} explains how our LLM backbone is used and trained to select skills.
        \item Section~\ref{app:prompts_for_HL} details our prompting strategy.
    \end{itemize}
    \item Section~\ref{app:low_level_policy} details the implementation of our low-level policy.
    \begin{itemize}
        \item Section~\ref{app:low_level_policy_architecture} explains the architecture of our lightweight neural networks used for the low-level policy.
        \item Section~\ref{app:low-level-training} details the low-level policy's training strategy based on AWR \cite{peng_advantageweightedregression2019}.
        \item Section~\ref{app:multigoal_buffer} presents a method to mitigate loss of plasticity and catastrophic forgetting during training of the low-level policy.
        \item Section~\ref{app:buffer_transitions_filtration} explains why it is necessary to filtrate a part of compiled trajectories and how our method do that.
    \end{itemize}
    \item Section~\ref{app:ll_sr_estimator} gives details about our low-level success rate estimator.
    \begin{itemize}
        \item Section~\ref{app:ll_success_estimator} explains how the low level success rate estimator is trained.
        \item Section~\ref{app:pg} gives the exact formula used to calculate the probability for adding an otion in the option set $\tilde{G}$ at step $k$.
        \item Section~\ref{app:prompt_ll_sr_estimator} shows the prompt given to our estimator.
        \item Section~\ref{app:training_of_LL_sr_estimator} provides details on the training strategy for the low-level success rate estimator.
        \item Section~\ref{app:i_tilde_composition} explains how the estimation is used to generate the set of skills the high-level policy samples from.
    \end{itemize}
    \item Section~\ref{app:magellan_implementation} describes how our autotelic goal sampler reimplements MAGELLAN \cite{gaven_magellan2025}.
    \begin{itemize}
        \item Section~\ref{app:magellan_adaptation} provides details on how we have adapted MAGELLAN to our problem.
        \item Section~\ref{app:training_of_HL_sr_estimator} shows the prompt given to MAGELLAN's success rate estimator.
    \end{itemize}
    \item Section~\ref{app:baselines_details} provides implementation details of our baselines.
    \begin{itemize}
        \item Section~\ref{app:FuN} describes our implementation of \textsc{FuN} \cite{vezhnevets_feudalnetworkshierarchicalreinforcement2017}.
        \item Section~\ref{app:poad_training} details our implementation of \textsc{poad}'s \cite{wen_reinforcing2024} training strategy.
        \item Section~\ref{app:boss_training} gives the practical implementation and training for the \textsc{BOSS}'s algorithm \cite{zhang_bootstrap2023}.
    \end{itemize}
    \item Section~\ref{app:hyperparameters} details the hyper-parameters we used.
    \item Section~\ref{app:compute_resources} gives the compute resources used for performing the experiments.
\end{itemize}

\newpage
\section{Environment detail} \label{app:env_detail}
\subsection{Achievement tree}
\label{app:achievement_tree}

We build upon the Crafter environment introduced by \citet{hafner_benchmarking2022}, introducing several modifications to emphasize learning in a heterogeneous and compositional goal space. Specifically, our environment features an achievement tree structure that explicitly defines the prerequisite relationships among goals, thereby encouraging agents to discover and exploit goal dependencies. The environment comprises $17$ distinct goals, categorized into three types: (i) $5$ movement-related goals requiring navigation to specific locations; (ii) $3$ collection-based goals that require prior construction of the appropriate tools —e.g., collecting stone and coal is only feasible after crafting the corresponding equipment; and (iii) $3$ crafting goals, which are achievable only when the agent possesses a sufficient quantity of the required resources in its inventory. This design introduces structured complexity that challenges agents to plan and act over extended temporal horizons.

Here is the list of all the goals and their requirements:

\begin{itemize}
    \item go to tree: facing a tree
    \item collect wood: facing a tree, no tool needed
    \item place table: facing grass, sand, path and having $2$ woods in inventory
    \item go to table: facing a table
    \item make wood pickaxe: facing a table and having $1$ wood in inventory
    \item go to stone: facing a stone
    \item collect stone: facing a stone, having a wood pickaxe
    \item go to coal: facing a coal
    \item collect coal: facing a coal, having a wood pickaxe
    \item place furnace: facing grass, sand, path and having $4$ stones in inventory
    \item go to furnace: facing a furnace
    \item make stone pickaxe: facing a table and having $1$ wood and $1$ stone in inventory
    \item go to iron: facing iron
    \item collect iron: facing iron, having a stone pickaxe
    \item make iron pickaxe: facing a furnace and having $1$ iron $1$ coal and $1$ wood in inventory
    \item go to diamond: face diamond
    \item collect diamond: face diamond with an iron pickaxe
\end{itemize}

\begin{figure}[ht]
    \centering
    \includegraphics[width=1\linewidth]{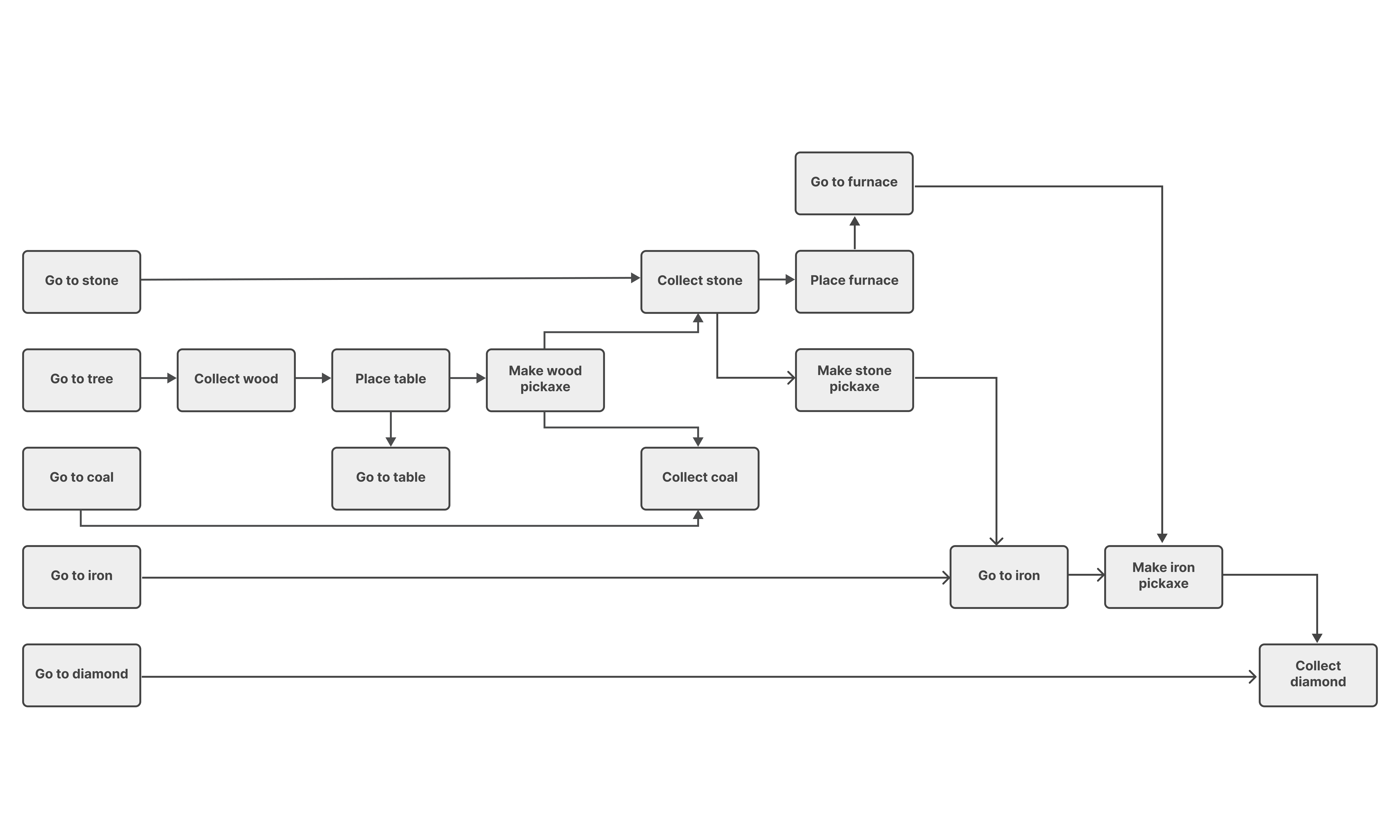}
    \caption{Achievement tree used in our experiments, the goal on the right necessitate some composition of the goals on the left.}
    \label{fig:tech_tree}
\end{figure}

\subsection{Modifications}
\label{app:modifications}

Similarly to \cite{du_guiding2023} we modify the original Crafter environment to isolate and emphasize the compositional aspects of goal-directed behaviour, removing elements primarily associated with survival dynamics. First, we eliminate hostile entities such as zombies and skeletons, as well as the need for the agent to eat or drink, thereby reducing perturbation-related distractions. Second, we convert the \textit{‘do’} action that originally represents several types of actions (meaning \textit{‘attack’} in front of a zombie but \textit{‘eat’} in front of a plant) into distinct actions (such as \textit{‘eat plant’} or \textit{‘attack zombie’}). %This disambiguates these elementary actions for LLMs. 
To improve the agent's navigational capabilities, we augment the goal space with \textit{"go to $X$"} objectives. These are always feasible for $X \in {\text{tree, stone, coal}}$, but may not be realizable for $X \in {\text{table, furnace}}$ depending on the current world state, introducing ambiguity that poses a challenge for the low-level success rate estimator (see Appendix~\ref{app:ll_sr_estimator}). Furthermore, we enrich the observation space by integrating a captioning module (see Appendix~\ref{app:observations}). The hierarchical policy operates under temporal constraints: the high-level controller is allowed up to $96$ steps (i.e., skill invocations), each of which may be executed by the low-level controller in up to $64$ steps. To enable multi-goal episodes while ensuring bounded episode length, we cap the environment at $155$ high-level steps before a reset occurs.

\subsection{Observations}
\label{app:observations}

The modified Crafter environment used in our experiments generates both a standard visual observation and a structured textual description. The visual observation consists of an RGB image of dimension $3 \times 32 \times 32$. The textual description acts as a sparse captioning of the visual input and is composed of multiple informative components, each providing structured context relevant to the agent’s decision-making. Specifically, the textual description includes:
\begin{itemize}
\item A general sentence describing the overall nature of the game.
\item The agent's current goal, along with the number of remaining high-level steps to achieve it.
\item The player's current position in the environment.
\item A description of all visible elements in the field of view, each identified by type and relative position using cardinal directions (e.g., "a tree is three steps north"). Each type of element is mentioned only once, even if multiple instances are visible.
\item A description of the block directly in front of the player.
\item A list of items currently in the player's inventory.
\item The positions of environment elements that have been placed by the player.
\item The available elementary actions.
\item If applicable, the list of admissible skills that can be invoked by the high-level policy.
\item The last action executed by the agent.
\end{itemize}
This detailed textual observation allows the high-level policy to operate with a semantically rich input space, facilitating more informed and context-aware decision-making. An example of such textual observation is given in Figure~\ref{fig:example_prompt_hl_1}, Figure~\ref{fig:example_prompt_hl_2}.

\newpage
\begin{figure}
    \centering
    \includegraphics[width=1\linewidth]{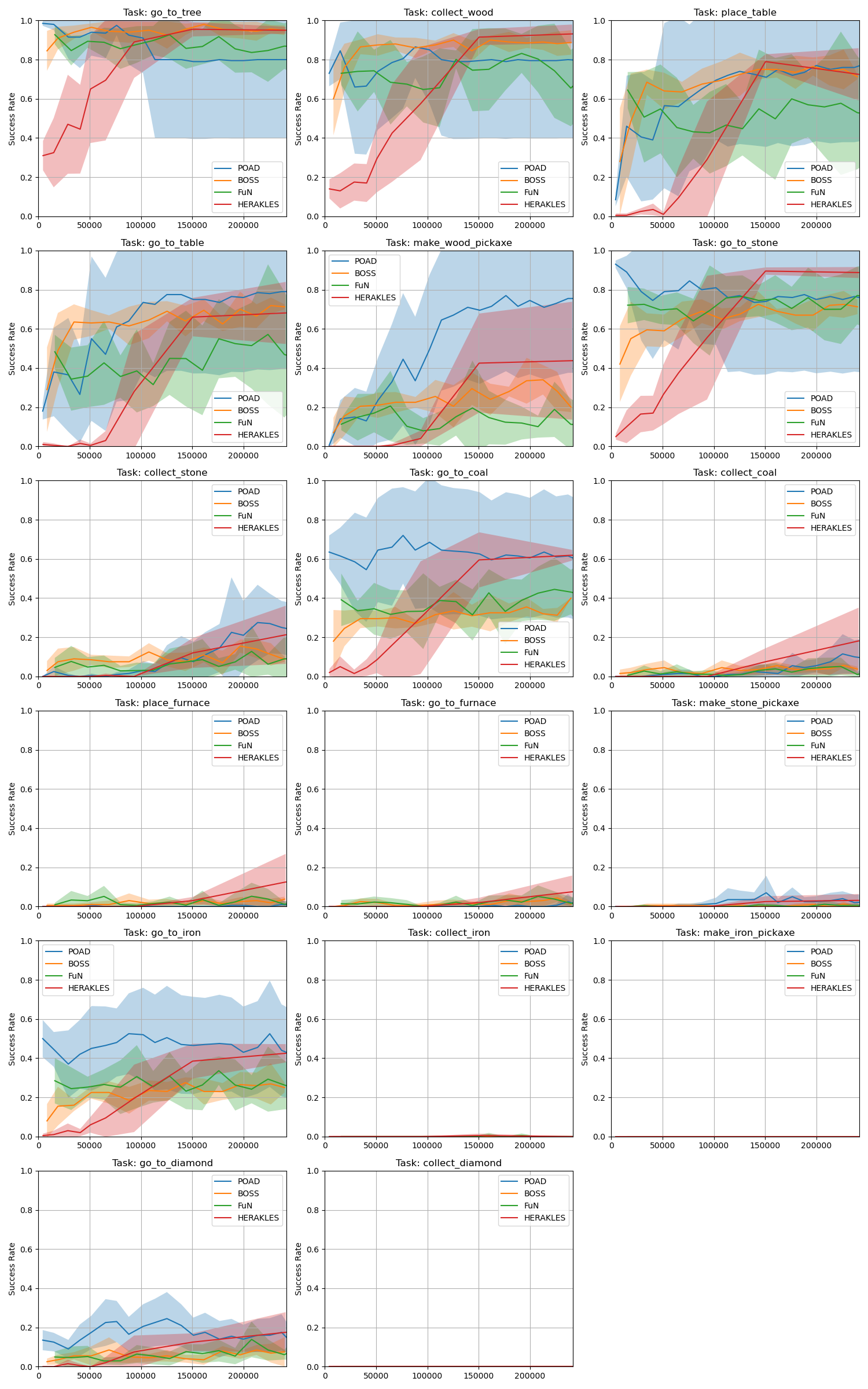}
    \caption{Success rate of each method per task depending on the number of interaction steps done in the environment. Shaded area represent standard deviation based on $5$ seeds.}
    \label{fig:full_sr_evolution_per_task_interaction_steps}
\end{figure}
\newpage

\section{Complementary Results}
\label{app:complementary_results}

\subsection{Evolution of success rate}
\label{app:evolution_of_success_rate}

In the Figure~\ref{fig:full_sr_evolution_per_task_interaction_steps}, we analyze the evolution of the success rate (SR) during training for each method. We measure the success rate every $28 000$ interaction steps by testing the agent on $40$ different held-out environments. For the test, the agent has to complete the goal in a maximum of $1000$ interaction steps with the environment and under $64$ high level steps. For the simple tasks such as \textit{"go to tree"} or \textit{place table} the baselines tend to perform better than \herakles as they do not need to learn a low level agent (\textsc{POAD}) or a classifier to propose the relevant subgoals to the high level policy (\textsc{FuN}, \textsc{BOSS}). However, for more complex goals such as \textit{collect coal} or \textit{place furnace} that requires long chain of actions, baselines tend to stagnates (\textsc{BOSS}) or are not able to learn in the given time (\textsc{POAD}). \herakles exhibits rapid improvement on more complex compositional tasks. For instance, the SR when \herakles started to master \textit{make wood pickaxe} around $150000$ steps it simultaneously starts to master \textit{collect stone} and \textit{place furnace} that reuses the \textit{make wood pickaxe}.  

%observe that \herakles consistently outperforms the baselines \textsc{poad} and \textsc{FuN} across all tasks and at every training stage. For simple navigation tasks such as \textit{"go to tree"}, both \textsc{poad} and \herakles achieve perfect performance from step $=0$. This suggests that \herakles' hierarchical agent initially relies on its high-level policy, which is identical to that of \textsc{poad} before the first update. Notably, \herakles exhibits rapid improvement on more complex compositional tasks. For instance, the SR for \textit{"place table"} increases from $0.5$ to $1.0$ in fewer than $5000$ steps once the prerequisite goal \textit{"collect wood"} is mastered. Similar patterns are observed for other tasks such as \textit{"make wood pickaxe"} and \textit{"collect stone"}, where mastery of foundational subgoals accelerates progress. In contrast, when a baseline method successfully learns an individual task, this progress does not translate to improvements on related downstream tasks, indicating a lack of effective compositional generalization.

\subsection{Generalization results}
\label{app:generalization_results}

\begin{table}
\centering
\begin{tabular}{c|cccc}
 & \herakles & \textsc{poad} & \textsc{FuN} &  \textsc{BOSS} \\ \hline
Original goals & 39.1 & 23.6 & 7.8 &  27.6\\
Synonyms & \begin{tabular}[c]{@{}c@{}}32.8\\ (\textbf{-16\%})\end{tabular} & \begin{tabular}[c]{@{}c@{}}18.2\\ (\textbf{-23\%})\end{tabular} & \begin{tabular}[c]{@{}c@{}}5.7\\ (\textbf{-26\%})\end{tabular} & \begin{tabular}[c]{@{}c@{}}24.8\\ (\textbf{-10\%})\end{tabular}
\end{tabular}
\caption{Generalization for synonyms goals.}
\label{tab:generalization_synonyms}
\end{table}

The ability to generalize to novel goals, particularly those expressed differently in natural language, is essential for open-ended agents to learn efficiently. In many cases, goals generated within a linguistic space may be semantically equivalent, either due to synonymy or because they correspond to behaviours already mastered under different phrasings.

\paragraph{Synonym generalization} We assess the generalization performance of \herakles, \textsc{poad}, and \textsc{FuN} on a set of synonym-based goals. For each original goal, such as \textit{"collect wood"}, we define a synonym set by selecting five alternative formulations (e.g., \textit{"gather wood"}, \textit{"harvest wood"}, \textit{"procure wood"}, \textit{"acquire wood"}, and \textit{"amass wood"}), and compute the average Crafter score across these variants. 
%To prevent bias from any single linguistic choice, the average success rate for a synonym category is computed as the mean success rate across all five synonymous formulations. 
Table~\ref{tab:synonyms} lists the complete set of synonym categories used in our evaluation. Of note, certain words such as \textit{"acquire"} appear across multiple categories (e.g., for both \textit{"collect"} and \textit{"make"}), requiring the agent to rely on contextual cues to disambiguate the intended task—a key competency in linguistically grounded goal spaces.

\begin{comment}
We quantify generalization performance using the average Crafter score: $$ \frac{1}{T} \Sigma_{t=0}^{T} S_c(t) dt,$$
where $T=30,000$ are the number of steps performed in the environment and where $S_c(t)$ denotes the crafter score at time $t$, defined as
$$S_c(t) = \exp( \frac{1}{N} \sum^N_{i=1} \ln(1 + \hat{sr}_i(t)))-1,$$ and $\hat{sr}_i(t)$ is the average success rate over the synonyms of goal $i$ at time $t$.
\end{comment}

We quantify generalization performance by extending the Crafter score for synonym reformulations as:
$$S^{syn}_c(t) = \exp( \frac{1}{|G|} \sum_{g \in G} \frac{1}{|Syn(g)|} \sum_{g' \in Syn(g)} \ln(1 + sr_{g'}(t)))-1,$$
where $Syn(g)$ stands for the set of possible reformulations of goal $g \in G$ given the list of synonyms given in Table \ref{tab:synonyms}, and $sr_{g'}(t)$ is the average success rate over a  reformulated goal $g' \in \bigcup_{g \in G} Syn(g)$ at step $t$ of the training. % synonyms of goal $i$ at time $t$.  
%This rate $sr_{g'}(t)$ is estimated by performing 8 runs using the model at step $t$, with the high-level policy conditioned for goal $g'$.  
%Figure~\ref{fig:generalization_synonym} plots $S^{syn}_c(t)$ every 2496 steps during policy training. %Table~\ref{tab:generalization_synonyms} aggregates these punctual measurements in a single averaged metric to render the rapidity of acquiring synonym generalization ability.  
The rate \( sr_{g'}(t) \) is estimated by performing 8 runs using the model at training step \( t \), with the high-level policy conditioned on the goal \( g' \).

%Figure~\ref{fig:generalization_synonym} shows the values of \( S^{\text{syn}}_c(t) \) recorded every 2,496 training steps, compared to the \( S_c(t) \) which uses original goals from $G$ only.  
%\( S^{\text{syn}}_c(t) \) is recorded every 2,496 and Table~\ref{tab:generalization_synonyms} aggregates the instantaneous measurements into a single averaged metric over the entire training period. %, in order to reflect the relative speeds at which the models acquire the ability to generalize to synonyms.

Table~\ref{tab:generalization_synonyms} contains the value of $ S^{\text{syn}}_c$ at the end of training. As illustrated in Table~\ref{tab:generalization_synonyms} \herakles preserves a high Crafter score in the synonym goal space, demonstrating a robust linguistic generalization. While \herakles experiences only a $16\%$ drop in average score relative to the original goal space, \textsc{poad}, \textsc{FuN} and \textsc{BOSS} exhibit more substantial decreases of $23\%$, $26\%$ and $10\%$, respectively. These results highlight the superiority of \herakles in handling semantic variability in goal specification.

\begin{table}[ht]
\centering
\caption{Synonyms used for the synonym generalization experiment.}
\begin{tabular}{r|ccc}
\multicolumn{1}{l|}{Original words} & \multicolumn{3}{c}{Synonyms} \\ \hline
\multirow{2}{*}{collect} & gather & acquire & procure \\
 & harvest & amass &  \\ \cline{2-4} 
\multirow{2}{*}{make} & craft & construct & build \\
 & acquire & create &  \\ \cline{2-4} 
\multirow{2}{*}{place} & put & putdown & install \\
 & deploy & position &  \\ \cline{2-4} 
\multirow{2}{*}{go} & move & walk & proceed \\
 & travel & run &
\end{tabular}
\label{tab:synonyms}
\end{table}

\paragraph{$n$-Compositionality}
A second form of generalization we investigate is \textit{$n$-compositionality}, which evaluates an agent's ability to scale learned behaviours to repeated instances of the same subgoal. In this setting, we construct new environments where each original goal must be achieved $n$ times consecutively, with $n \in {2, 3, 4}$. For example, the original goal \textit{"collect wood"} is reformulated as \textit{"collect 2 woods"} in the $2$-compositional environment. This setup tests the compositional depth of the agent’s policy and its capacity to generalize beyond one-shot achievement.

We exclude \textit{"go to $X$"} goals from the $n$-compositional environment, as repeating these goals lacks meaningful interpretation—one cannot meaningfully go to the same object $n$ times in succession.

Table~\ref{tab:generalization_n_compositional} illustrates that the Crafter score decreases for all methods as $n$ increases, reflecting the growing complexity of the environment. However, \herakles exhibits a significantly smaller relative performance drop compared to the baselines. For instance, at $n=4$, \herakles incurs a $51\%$ reduction in Crafter score, whereas \textsc{poad} suffers a $92\%$ drop, \textsc{FuN} a $82\%$ and \textsc{BOSS} $64\%$. These results indicate that \herakles is better equipped to generalize through compositional repetition, a key property for agents operating in scalable, open-ended settings.

When analyzing performance across both generalization settings—synonym-based and $n$-compositional—we observe complementary strengths in the two baseline methods. Although \textsc{poad} exhibits a lower overall average Crafter score, it outperforms \textsc{FuN} on synonym generalization tasks. This is likely attributable to the linguistic capabilities embedded in its large language model (LLM) backbone, in contrast to \textsc{FuN}, which relies solely on a static embedding for goal representation. Conversely, \textsc{FuN}'s hierarchical architecture enables stronger performance in the $n$-compositional setting, where the reuse and chaining of subskills are crucial. Notably, \herakles, which integrates the strengths of both approaches—leveraging both a linguistically grounded high-level policy and a hierarchical skill structure—consistently outperforms both baselines across all generalization regimes.

\begin{table}
\centering
\caption{Generalization for $n$-compositional goals.}
\begin{tabular}{c|cccc}
 & \herakles & \textsc{poad} & \textsc{FuN} &  \textsc{BOSS} \\ \hline
Original goals & 39.1 & 23.6 & 7.8 &  27.6\\
2 Compositional & \begin{tabular}[c]{@{}c@{}}33.2\\ (\textbf{-15\%})\end{tabular} & \begin{tabular}[c]{@{}c@{}}0.396\\ (\textbf{-68\%})\end{tabular} & \begin{tabular}[c]{@{}c@{}}5.11\\ (\textbf{-34\%})\end{tabular} & \begin{tabular}[c]{@{}c@{}}18.8\\ (\textbf{-32\%})\end{tabular} \\
3 Compositional & \begin{tabular}[c]{@{}c@{}}26.6\\ (\textbf{-32\%})\end{tabular} & \begin{tabular}[c]{@{}c@{}}0.231\\ (\textbf{-82\%})\end{tabular} & \begin{tabular}[c]{@{}c@{}}2.51\\ (\textbf{-67\%})\end{tabular} & \begin{tabular}[c]{@{}c@{}}11.6\\ (\textbf{-58\%})\end{tabular} \\
\multirow{2}{*}{4 Compositional} & \multirow{2}{*}{\begin{tabular}[c]{@{}c@{}}19.2\\ (\textbf{-51\%})\end{tabular}} & \multirow{2}{*}{\begin{tabular}[c]{@{}c@{}}0.102\\ (\textbf{-92\%})\end{tabular}} & \multirow{2}{*}{\begin{tabular}[c]{@{}c@{}}1.40\\ (\textbf{-82\%})\end{tabular}} & \multirow{2}{*}{\begin{tabular}[c]{@{}c@{}}9.94\\ (\textbf{-64\%})\end{tabular}} \\
 &  &  & 
\end{tabular}
\label{tab:generalization_n_compositional}
\end{table}

\subsection{Evolution of high-level sampling strategy}
\label{app:evolution_hl_sampling_strategy}

\begin{figure}[!ht]
  \centering
  \begin{subfigure}{\textwidth}
    \centering
    \includegraphics[width=0.6\linewidth]{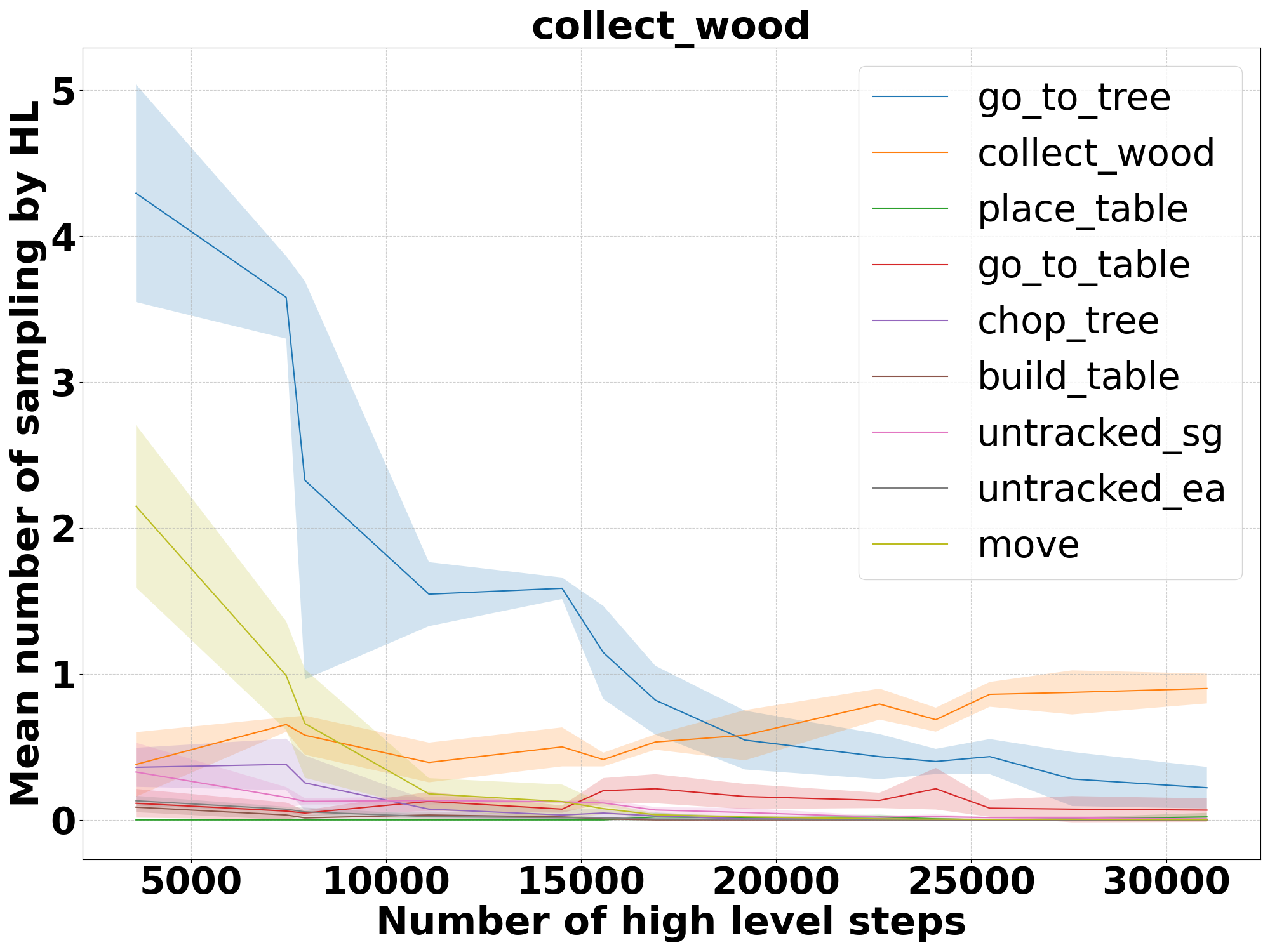}
    \caption{Collect wood.}
    \label{fig:collect_wood}
  \end{subfigure}
  \begin{subfigure}{\textwidth}
    \centering
    \includegraphics[width=0.6\linewidth]{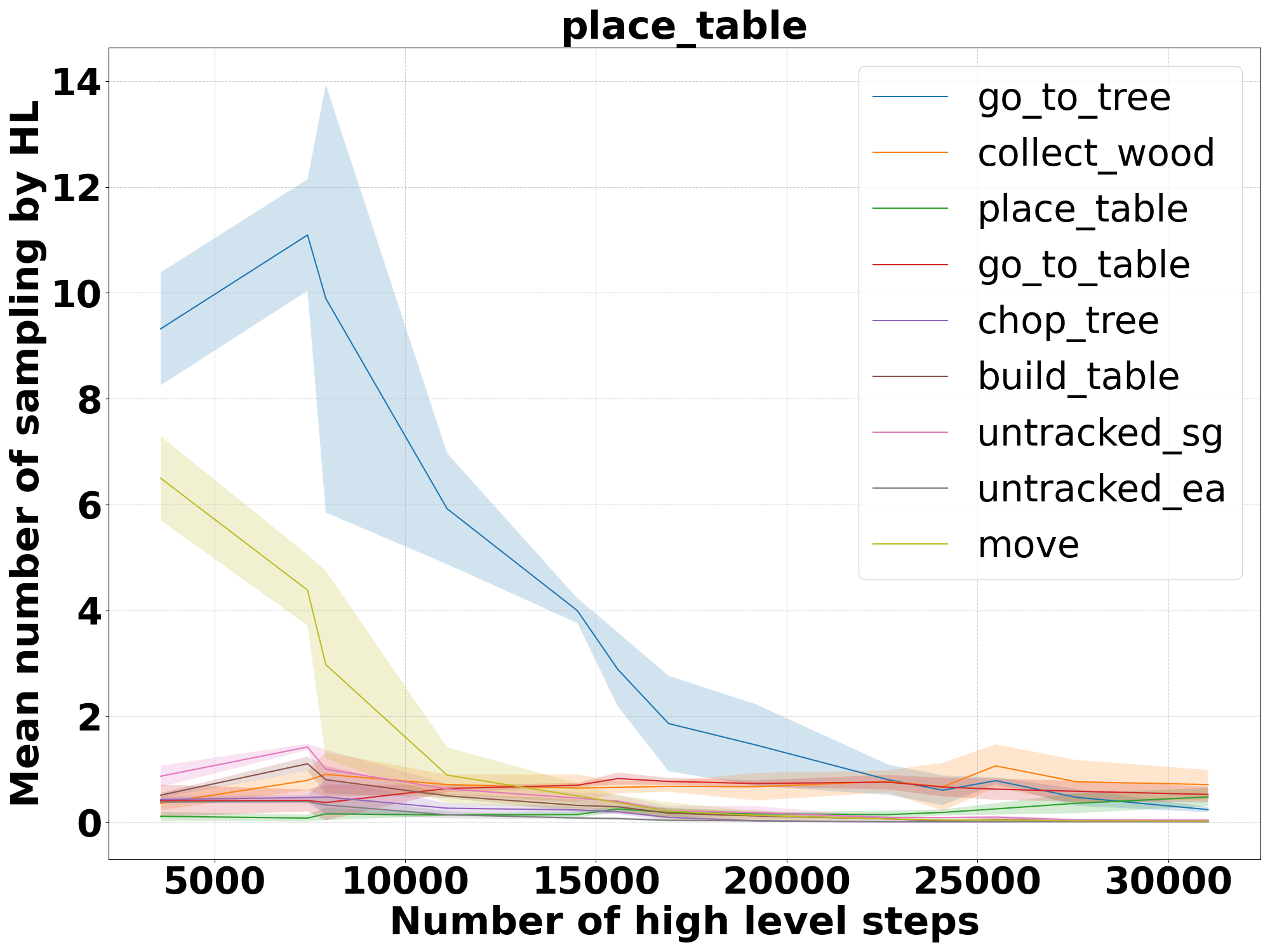}
    \caption{Place table.}
    \label{fig:place_table}
  \end{subfigure}
  \begin{subfigure}{\textwidth}
    \centering
    \includegraphics[width=0.6\linewidth]{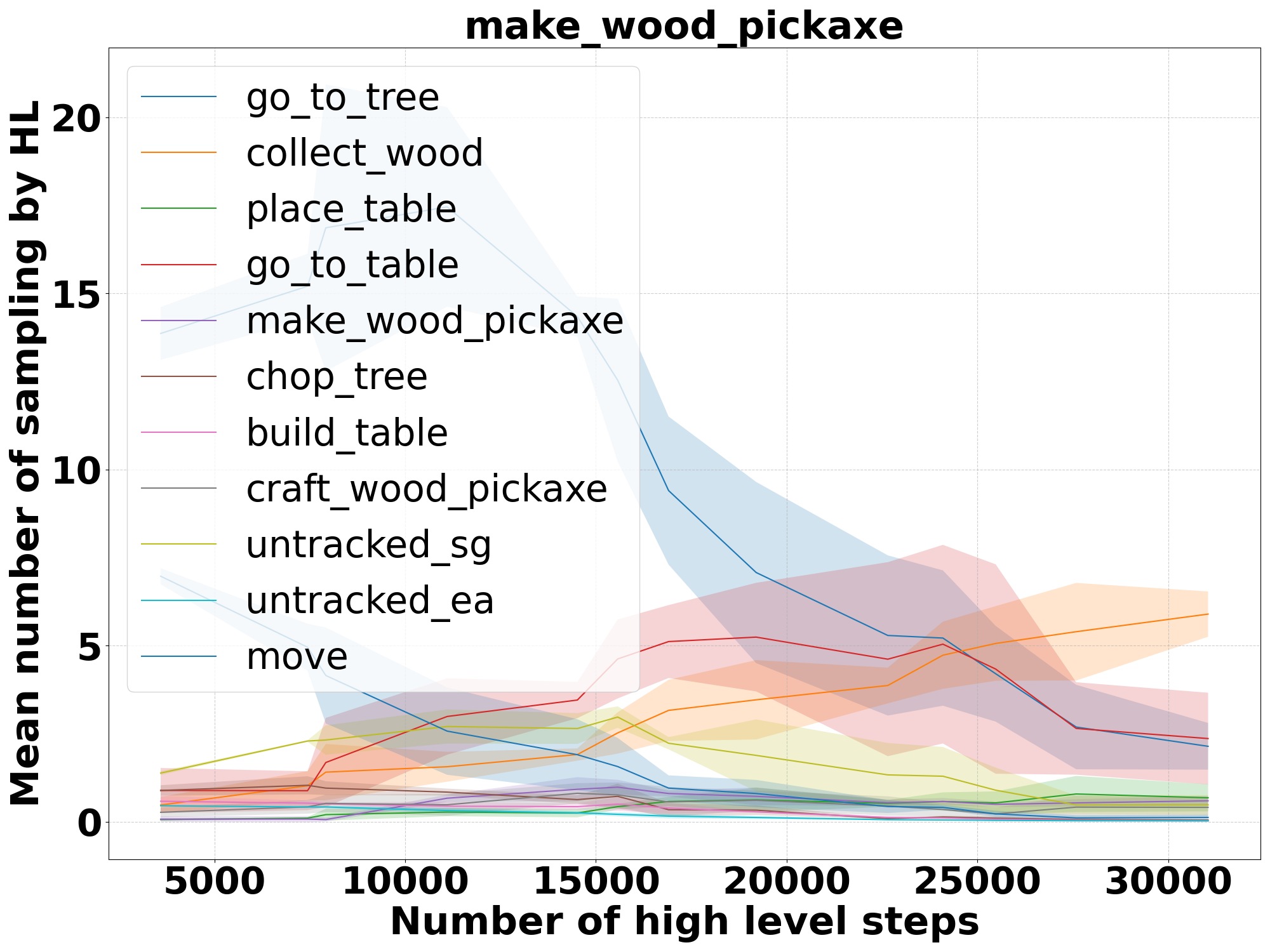}
    \caption{Make wood pickaxe.}
    \label{fig:make_wood_pickaxe}
  \end{subfigure}
  \caption{Evolution of sampling strategy of high level.}
  \label{fig:hl_sampling_strategy}
\end{figure}

\herakles leverages the continuous compilation of mastered goals into low-level skills, gradually transferring control from the hierarchical agent to the low-level controller. In this section, we analyze how this mechanism shapes the high-level policy's sampling strategy throughout training. According to the learning dynamics described in Section~\ref{methods}, we hypothesize a three-phase progression: initially, the high-level policy broadly samples elementary actions to guide the undertrained low-level policy; subsequently, as the low-level improves, skill calls become more frequent; and finally, the high-level converges to invoking the compiled skill corresponding to the goal directly, relying fully on the low-level policy for execution. For instance, for the goal "collect wood", the high level would sample the \textit{"collect wood"} skill.

In Figure~\ref{fig:hl_sampling_strategy}, we report the mean number of skill or action calls per trajectory. For clarity, we aggregate all untracked compiled skills and untracked elementary actions into \textit{untracked\_sg} and \textit{untracked\_ea}, respectively, and average all \textit{"move"} actions under a single \textit{"move"} category.

Across the three subplots of Figure~\ref{fig:hl_sampling_strategy}, we observe that before $10{,}000$ high-level steps, the policy predominantly samples elementary actions of type \textit{"move"} and the skill \textit{"go to tree"}. Since gathering wood is a prerequisite for nearly all goals, this early focus is expected. Notably, the skill \textit{"go to tree"} is oversampled—approximately $10$ times per trajectory—because the low-level policy has yet to master it, despite the skill being highly accessible (random agents succeed $90\%$ of the time in $128$ steps).

As shown in Figure~\ref{fig:collect_wood}, the elementary action \textit{"chop tree"} is also sampled about $0.5$ times per trajectory in the same early phase. Between $10{,}000$ and $20{,}000$ steps, elementary action calls decrease and the high-level policy increasingly delegates to compiled skills. This shift is evident in Figure~\ref{fig:make_wood_pickaxe}, where \textit{"collect wood"} and \textit{"go to table"} are more frequently invoked.

By $25{,}000$ steps, the high-level policy either directly selects compiled skills, as in Figure~\ref{fig:collect_wood} with \textit{"collect wood"}, or replaces simpler skills with more complex ones, as seen in Figures~\ref{fig:place_table} and \ref{fig:make_wood_pickaxe}, where \textit{"collect wood"} replaces the earlier calls to \textit{"go to tree"}.

These observations confirm that \herakles adheres to the hypothesized learning dynamics, effectively leveraging the hierarchy to bootstrap new skills from previously acquired ones, thereby enabling efficient goal acquisition over time.

\section{Hierarchical Agent}
\label{app:hierarchical_agent}
\subsection{HERAKLES algorithm}
\label{app:herakles_algorithm}

In this section, we provide the pseudocode of the \herakles algorithm (see Algorithm~\ref{alg:herakles}), showing the core training loop and its hierarchical decision-making structure. For clarity and brevity, we omit the update mechanisms for the MAGELLAN goal-sampling module \cite{gaven_magellan2025} detail in Appendix~\ref{app:magellan_implementation}.

\begin{algorithm}
\caption{HERAKLES}
\label{alg:herakles}
\begin{algorithmic}
\Require $\pihl$ the high-level policy, $\pill$ the low-level policy, $M$ the MAGELLAN module \cite{gaven_magellan2025}, $G$ the goal space to be learned, $\mathcal{A}$ set of low level sensory motor commands, $\mathcal{B_{\text{HL}}}$ $\pihl$ buffer, $\mathcal{B^{\text{h}}_{\text{LL}}}$ $\pill$ buffer for trajectories collected from the hierarchical policy, $\mathcal{B^{\text{sg}}_{\text{LL}}}$ $\pill$ buffer for trajectories collected from $\pill$ policy alone executing a goal decided by $\pihl$, $ C_{\theta_k}$ the $\pill$ competency estimator and $env$ the environment.
\State 
\State $(obs^{visual}, obs^{textual}) \gets env.reset()$ \Comment{see Appendix~\ref{app:observations}}
\State $g \gets M(obs^{textual})$ \Comment{Choose a goal $g$ in an autotelic way}
\State $done_g, r^g \gets env.verify(g)$ \Comment{verify checks if $g$ is done, $r^{g}$ reward for $g$}
\While{$done_g$ not $\texttt{True}$}
\State $\tilde{G} \gets \{ \}$ \Comment{Instantiate the skill set}
\For{$g' \in G$} \Comment{Adding a skill with probability $p_{g'}$}
\State $\tilde{G} \gets\tilde{G} \cup \{g'\} \sim p_{g'}$ \Comment{$p_{g'}$ expression in Appendix~\ref{app:pg}}
\EndFor
\State $\tilde{g} \gets \pi^{HL}(g, (obs^{textual},\tilde{G}))$
\State $done_{g}, r^{g} \gets env.verify(g)$ 
\State $done_{\tilde{g}}, r^{\tilde{g}} \gets env.verify(\tilde{g})$
\While{$ done_{\tilde{g}}$ not $\texttt{True}$ and $ done_{g}$ not $\texttt{True}$} \Comment{$\pill$ executes the skill in the environment}
\State $ a \gets \pi^{LL}(\tilde{g},  obs^{visual})$ \Comment{$a \in \mathcal{A}$}
\State $(obs^{visual}_{next}, obs^{textual}_{next}) \gets env.step(a)$
\State $done_{g}, r^{g} \gets env.verify(g)$ 
\State $done_{\tilde{g}}, r^{\tilde{g}} \gets env.verify(\tilde{g})$
\State $\mathcal{B^{\text{sg}}_{\text{LL}}} \gets (\tilde{g}, obs^{visual}, a, r^{\tilde{g}})$
\Comment{Transition seen as targeting $\tilde{g}$}
\State $\mathcal{B^{\text{h}}_{\text{LL}}} \gets (g, obs^{visual}, a, r^{g})$ \Comment{Transition seen as targeting $g$}
\State $obs^{visual} \gets obs^{visual}_{next}$
\If{enough new transitions in $\mathcal{B_{\text{LL}}}$}
\State Update $\pill$ on $\mathcal{B^{\text{h}}_{\text{LL}}} \cup \mathcal{B^{\text{sg}}_{\text{LL}}}$ \Comment{Using AWR \cite{peng_advantageweightedregression2019}}
\State Update $ C_{\theta_k}$ using $\mathcal{B^{\text{sg}}_{\text{LL}}}$ \Comment{Appendix~\ref{app:ll_sr_estimator}}
\EndIf
\If{$done_{\tilde{g}}$ or $ done_{g}$}
\State $\mathcal{B_{\text{HL}}} \gets (g, (obs^{textual},\tilde{G}), a, r^{g})$
\EndIf
\EndWhile
\State Update $\pihl$ $\mathcal{B_{\text{HL}}}$ \Comment{Using \textsc{poad} \cite{wen_reinforcing2024}}
\State $obs^{textual} \gets obs^{textual}_{next}$
\EndWhile
\end{algorithmic}
\end{algorithm}

\subsection{HERAKLES implementation}
\label{app:herakles_implementation}

\begin{figure}
    \centering
    \includegraphics[width=0.85\linewidth]{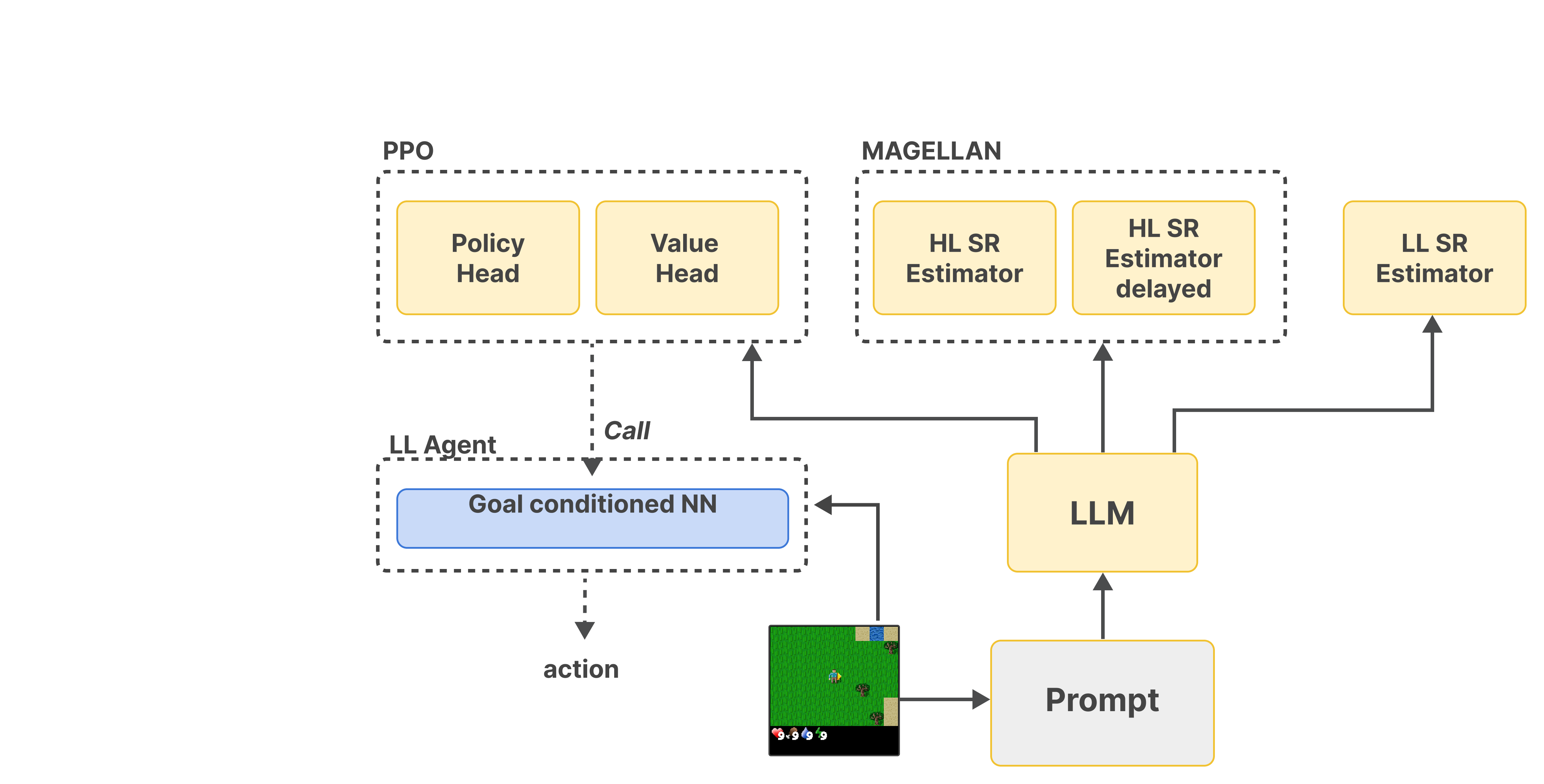}
    \caption{HERAKLES architecture. In yellow are the parts that useda LLM backbone and are trained with LoRA adapters~\cite{Hu_LoRA2021}. In blue is the simple goal conditioned neural network that represent the low level policy, more details on its architecture in Appendix~\ref{app:low_level_policy_architecture}.}
    \label{fig:herakles_architecture}
\end{figure}

To instantiate the hierarchical agent described in Section~\ref{methods}, we employ a large language model (LLM)-backboned agent as the high-level policy (detailed in Appendix~\ref{app:high_level_policy}) and a goal conditioned lightweight neural networks as the low-level policy (see Appendix~\ref{app:low_level_policy}). The Figure~\ref{fig:herakles_architecture} gives an overview of the overall architecture.

In accordance with the formulation in Section~\ref{methods}, we assume that $\forall (a,s) \in A \times S, ; \pi^{LL}(a|s, a) = 1$. Consequently, when the high-level policy selects a skill that corresponds to an elementary action—e.g., \textit{"move right"}—we bypass the low-level controller and directly execute the action in the environment.

\herakles further incorporates a low-level success rate estimator to manage skill inclusion in the filtered skill set $\tilde{G}$, and leverages MAGELLAN to guide goal sampling. These components are described in Sections~\ref{app:ll_sr_estimator} and~\ref{app:magellan_implementation}, respectively.

Each module—the high-level policy, the individual low-level networks, the success rate estimator, and MAGELLAN—is trained independently using separate Adam optimizers. Detailed hyper-parameter configurations for each optimizer are provided in Appendix~\ref{app:hyperparameters}.

To help reproducibility, we make our code accessible at: \url{https://github.com/AnonymousUser530/HERAKLES}.

\section{High-level policy}
\label{app:high_level_policy}
\subsection{High-level policy architecture}
\label{app:high_level_policy_architecture}

The high-level policy architecture in \herakles builds upon GLAM~\cite{carta_grounding2023}, which employs an actor-critic framework grounded in a large language model (LLM) backbone. The value function is implemented as a two-layer multilayer perceptron (MLP) with $1024$ sigmoid-activated units, applied to the final hidden state of the decoder. Both the policy and the value head utilize shared LoRA adapters~\cite{Hu_LoRA2021}, enabling parameter-efficient fine-tuning of the LLM.

In contrast to GLAM, skill selection in \herakles is achieved through constrained decoding over the set of admissible skills $\tilde{G}$. Instead of sampling tokens from the entire vocabulary, token generation is restricted to sequences that correspond to valid skills in $\tilde{G}$, ensuring both syntactic validity and semantic precision.

Training of the high-level policy is conducted using \textsc{poad}~\cite{wen_reinforcing2024}, an algorithm that adapts PPO to operate at the token level for language agents. The policy is updated every $2496$ high-level transitions, using \textsc{poad} with $4$ training epochs per update cycle. In addition to the \textsc{poad} loss we add a penalty term based on Kullback-Liebler divergence:
$$\beta_{KL}KL(\pi^{HL}, \pi^{HL}_{original}),$$
with $\beta_{KL}$ an hyper-parameter controlling for the importance of this penalty term. It ensures that the high-level policy does not drift too much from the original policy and retains some plasticity, allowing it to adapt to the apparition of new usable skills during training. 

\subsection{Prompts for the High-Level policy}
\label{app:prompts_for_HL}
In this section, we provide an illustrative example of a prompt given to the high-level policy in \herakles. In the depicted scenario, the environment has not been reset since the previous interaction, a new goal, \textit{"make furnace"} has been sampled and the agent has already done one high level step. The prompt includes a description of the current state, the previously executed skill, and the list of admissible skills $\tilde{G}$. This setup allows the high-level policy to condition its next action on both the recent interaction history and the updated goal. Such prompt structure ensures that the LLM-based policy can contextualize its decision-making process in a temporally coherent and semantically grounded manner.
\begin{figure}[!ht]
\centering
\begin{adjustbox}{max width=\textwidth}
\begin{tcolorbox}[colback=gray!5!white, colframe=gray!75!black, title=Example Prompt for the high-level policy (1/2)]
\small
\ttfamily
You are playing a Minecraft like game. You can use elementary actions or, if available, more efficient low-level policies.

Your task: Make to furnace in 64.0 steps

You have already done 1 step.

Your coordinates: (29,10)

You see:
\begin{itemize}
    \item grass 1 step to your west
    \item path 5 steps to your south-west
    \item sand 4 steps to your south-east
    \item tree 3 steps to your east
    \item table 1 step to your north
    \item plant 6 steps to your south-west
\end{itemize}

You face table at your front.

Your inventory:
\begin{itemize}
    \item sapling: 9
    \item wood: 4
    \item coal: 2
    \item wood pickaxe: 4
\end{itemize}

You placed table at (4,3)
You placed table at (23,13)
You placed table at (29,9)
You placed plant at (0,12)
You placed plant at (0,0)
You placed stone at (13,0)

Elementary actions you can take:
\begin{itemize}
    \item move left
    \item move right
    \item move up
    \item move down
    \item sleep
    \item consume cow
    \item consume plant
    \item attack zombie
    \item attack skeleton
    \item attack cow
    \item chop tree (require facing tree)
    \item chop bush (require facing bush)
    
\end{itemize}
\end{tcolorbox}
\end{adjustbox}
\caption{Example of a prompt given to the agent high-level policy (1/2).}
\label{fig:example_prompt_hl_1}
\end{figure}
\newpage
\begin{figure}[!ht]
\centering
\begin{adjustbox}{max width=\textwidth}
\begin{tcolorbox}[colback=gray!5!white, colframe=gray!75!black, title=Example Prompt for the high-level policy (2/2)]
\small
\ttfamily
\begin{itemize}
    \item chop grass (require facing grass)
    \item extract stone (require 1 wood pickaxe in your inventory while facing stone)
    \item extract coal (require 1 wood pickaxe in your inventory while facing coal)
    \item extract iron (require 1 stone pickaxe in your inventory while facing iron)
    \item extract diamond (require 1 iron pickaxe in your inventory while facing diamond)
    \item drink water
    \item put stone (require 1 stone in your inventory)
    \item build table (require 2 woods in your inventory)
    \item build furnace (require 4 stones in your inventory)
    \item put plant (require 1 sapling in your inventory)
    \item craft wood pickaxe (require 1 wood in your inventory while facing a table)
    \item craft stone pickaxe (require 1 wood, 1 stone in your inventory while facing a table)
    \item craft iron pickaxe (require 1 wood, 1 coal, 1 iron in your inventory while facing a furnace)
    \item craft wood sword (require 1 wood in your inventory while facing a table)
    \item craft stone sword (require 1 wood, 1 stone in your inventory while facing a table)
    \item craft iron sword (require 1 wood, 1 coal, 1 iron in your inventory while facing a furnace)
\end{itemize}

Low-level policies you can call:
\begin{itemize}
    \item go to tree
    \item collect wood
    \item place table
    \item go to table
\end{itemize}

The last action you took: collect wood

Your action: 
\end{tcolorbox}
\end{adjustbox}
\caption{Example of a prompt given to the agent high-level policy (2/2)
.}
\label{fig:example_prompt_hl_2}
\end{figure}

\newpage
\section{Low-level policy}
\label{app:low_level_policy}
\subsection{Low-level policy architecture}
\label{app:low_level_policy_architecture}

\begin{figure}
    \centering
        \includegraphics[width=0.8\linewidth]{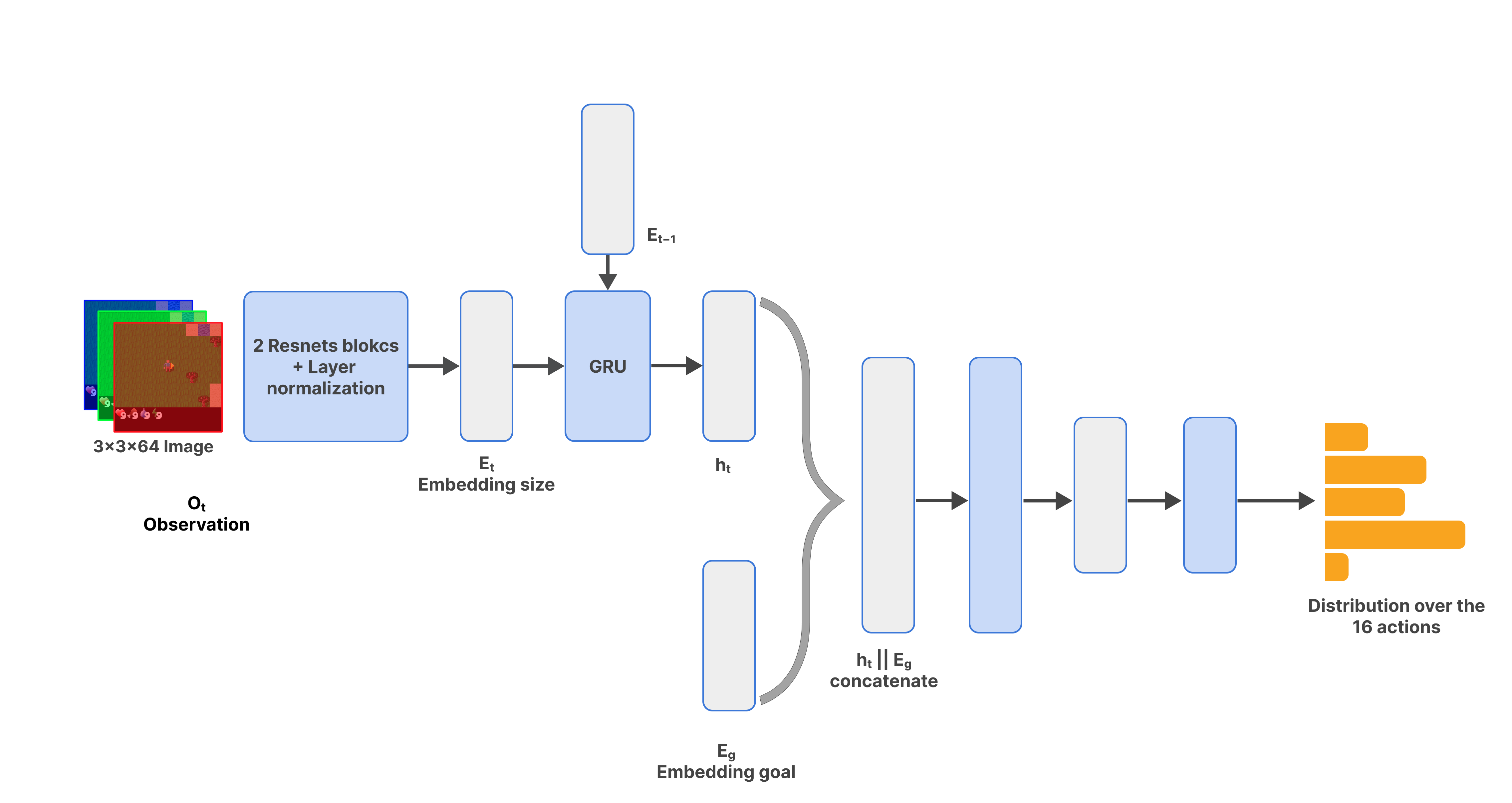}
    \caption{Low-level policy architecture. The blue rectangles represent neural network layers and the grey rectangles the outputted embeddings.}
    \label{fig:low_level_policy_architecture}
\end{figure}

The low-level policy in \herakles is designed to be lightweight compared to the high-level language model and operates directly on raw sensory input, specifically RGB observations from the environment. To extract features and generate elementary actions, we adapt the architecture proposed by \citet{moon_discovering2023}. This model employs a two-block ResNet~\cite{he2015deepresiduallearningimage} with channel dimensions $[64, 128, 128]$ to encode visual information. The resulting feature map is flattened and passed through a GRU layer \cite{cho_learning2014} to build a history of the trajectory embedded in $h_t$. $h_t$ is concatenated with the embedding of the goal $E_g$ and passes through two fully connected layers of sizes $1024$ and $128$, respectively. The final embedding is then fed into a sigmoid-activated output layer of dimension $16$ to produce a probability distribution in the discrete space of elementary actions. A low-level policy architecture diagram is shown in Figure~\ref{fig:low_level_policy_architecture}. The critic architecture is similar to the actor but the last layer has an output dimension of $1$. The low-level agent contains approximately $2$ million parameters, which constitutes only $0.03\%$ of the total parameter count of the high-level policy, ensuring computational efficiency.

\subsection{Low-level training}
\label{app:low-level-training}

We reimplement the Advantage-Weighted Regression (AWR) algorithm~\cite{peng_advantageweightedregression2019}
to train the low-level policy in \herakles. AWR is selected for its simplicity and robustness, as well as
its flexibility in integrating heterogeneous data sources into a unified replay buffer. Specifically, we
leverage two types of transitions: \textit{(i)} those collected from the low-level policy executing a given
skill, enabling continual skill refinement, and \textit{(ii)} transitions sampled from successful
trajectories of the hierarchical agent, used to distil composite behaviours into the low-level policy.

This dual sourcing enables multiplicative relabelling, analogous to Hindsight Experience Replay
(HER)~\cite{andrychowicz_hindsight2017}. For instance, a transition such as \textit{"facing a tree and
chopping wood"} can simultaneously reinforce the skill \textit{"collect wood"} and contribute to the
higher-level objective \textit{"place table"}. This mechanism allows the low-level agent to acquire new
skills by imitating hierarchical behaviours while refining its competence on previously mastered ones.

We employ sparse binary rewards to signal skill completion and goal achievement. For failed
trajectories, i.e., when a skill is not completed within the $64$-step budget, we substitute the
zero reward with the critic's value estimate as a proxy signal. This effectively smooths the
discontinuity introduced by the fixed time horizon and provides a more informative learning signal.

Determining \emph{when} to update the low-level policy requires balancing sample efficiency with
overall training throughput. As the agent progresses, it gains access to an increasingly large set of
subgoals, each generating a growing stream of transitions. Naively updating the low-level policy at a
fixed transition interval would cause it to over-train on frequently visited, elementary goals, such
as \textit{go to wood}, which is invoked as a prerequisite for nearly every higher-level objective
like \textit{place table}, while starving the system of new experience. Conversely, updating too
infrequently wastes the information carried by newly collected transitions.

To reconcile these competing pressures, we condition each update on three criteria: \textit{(i)} a
minimum number of new transitions must have accumulated since the last update, ensuring a sufficiently
large and fresh batch; \textit{(ii)} the goal must have a non-trivial success rate in the buffer,
filtering out goals for which the positive signal is too sparse to provide a reliable learning
gradient; and \textit{(iii)} the update probability decays with the number of low-level updates
performed since the last high-level update, preventing runaway specialisation on easy goals.
Formally, the update probability for goal $g$ is:

\begin{equation}
    p_{\text{update}}(g) \;=\;
    \mathbf{1}_{\Delta n(g)\,>\,2496}
    \;\cdot\;
    \mathbf{1}_{SR_{\tau}(g)\,>\,\alpha}
    \;\cdot\;
    \bigl(1 - SR_{\tau}(g)\bigr)^{N_{\text{ll}}(g)},
    \label{eq:ll_update_prob}
\end{equation}

where:
\begin{itemize}
    \item $\Delta n(g)$ denotes the number of new transitions for goal $g$ added to the replay buffer $\mathcal{B}_{LL}(g)$ since its last update;
    \item $SR_{\tau}(g)$ is the fraction of successful trajectories for goal $g$ currently stored in the buffer $\mathcal{B}_{LL}(g)$ with $\alpha=0.1$; and
    \item $N_{\text{ll}}(g)$ is the number of low-level updates performed on goal $g$ since the most recent high-level policy update.
\end{itemize}
The second indicator suppresses updates for goals that are too difficult and lack sufficient positive
examples, avoiding noisy gradient estimates. The exponential decay in the third term ensures that
goals which are mastered early, such as \textit{go to tree}, are not over-trained at the expense
of broader exploration.

\subsection{Multigoal buffer composition}
\label{app:multigoal_buffer}

\begin{figure}
    \centering
        \includegraphics[width=0.7\linewidth]{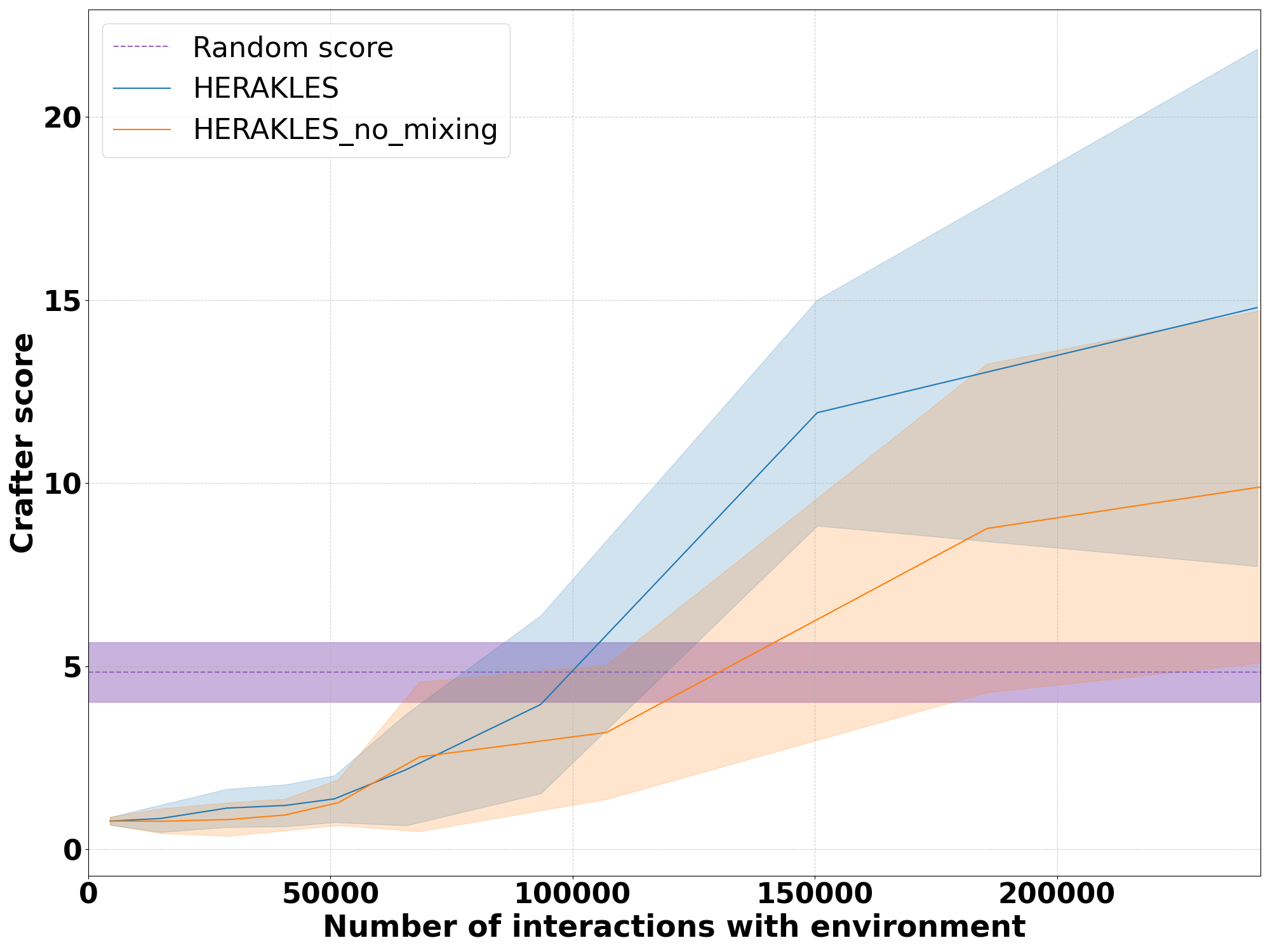}
    \caption{Evolution of the crafter score with and without doing the update of $\pill$ with a mixture of goal as regularisation.}
    \label{fig:ablation_awr_buffer_mixing_goal}
\end{figure}

In our method, $\pill$ is trained to achieve multiple goals in a continual learning setting, where new goals are introduced progressively during training. To mitigate loss of plasticity and catastrophic forgetting, we augment the training process as follows: whenever $\pill$ is updated using the replay buffer $\mathcal{B}_{LL}(g_i)$ associated with a goal $g_i$ we include up to a fraction $\beta_{sample}$ of transitions uniformly sampled from buffers corresponding to other goals $g_j$ with $i \neq j$. In our experiments, we set $\beta_{sample}=0.2$. The idea is that $\pill$ would not optimize only for $g_i$ forgetting its previous trainings on other goals.

In Figure~\ref{fig:ablation_awr_buffer_mixing_goal} we propose an ablation where there is no mixing of goals in the updates of $\pill$ during training. We observe that without such mixing the overall performances decrease strongly with a score of $10$ when there is no mixing instead of $15$ after $250000$ steps in the environment. Regularising the $\pill$ updates by adding a mixt of transitions from other goals help the low level to learn in a more robust manner.

\subsection{Buffer transitions filtration}
\label{app:buffer_transitions_filtration}

When the hierarchical agent attempts a new goal ( $g'$ ), the high-level policy explores by sampling a large number of options. The resulting relabelled trajectories added to the replay buffer of $\pill$ for ( $g'$ ) tend to be extremely long in terms of primitive actions—often several times longer than ( $N^{LL}$ ), the maximum number of steps allowed for $\pill$. Moreover, these trajectories are typically unrewarded, generating substantial noise during the early stages of training $\pill$ on ( $g'$ ).

To mitigate this issue, we prune a subset of these failed trajectories based on the hierarchical agent’s learning progress, denoted ( $LP^{hierarchical}$ ). Specifically, we retain a compiled failed trajectory only if its length ( $l_{compiled\_failed}$ ) satisfies
$$
l_{compiled\_failed}
< \frac{1}{1 - LP^{\text{hierarchical}}} , N^{LL}.
$$

We focus particularly on failed trajectories when the competency of the hierarchical policy is evolving, as such failures are more likely to carry meaningful learning signals.

\section{Low-level success rate estimator}
\label{app:ll_sr_estimator}

\subsection{Low-level success rate estimator training}
\label{app:ll_success_estimator}

To measure the probability of success $\mathbb{E}_{\pi^{LL}(.|g), s_k \sim D_{\pi^{LL}}}[r^{g}| O(s_k)]$, we leverage the LLM used by $\pihl$ by learning the parameters $\theta_k$ of a competence estimator $C_{\theta_k}(s_k, g) \approx \mathbb{E}_{\pi^{LL}(.|g), s_k \sim D_{\pi^{LL}}}[r^{g}| O(s_k)]$, following a strategy similar to that of \cite{gaven_magellan2025}. We pass in the LLM a prompt describing the pair $(s, g) \in S \times G$ which produces a latent representation on top of its final decoder block for the last token. The representation is passed to a multilayer perceptron (MLP) that outputs the estimated success probability $C_{\theta_k}(s, g)$. We train both the
LLM and the MLP by minimizing the binary cross-entropy: $\mathcal{L}(\theta_t) = \mathbb{E}_{((s, g), r) \sim \mathcal{D}_t} \left[ BCE(r
, C_{\theta_t}(s, g)) \right]$, with $\mathcal{D}_t$ a buffer containing the most recent $((s, g), r)$ pairs. In practice, we train two separate versions of the same initial LLM using LoRA adapters \cite{Hu_LoRA2021}: one for $\pihl$ and one for our competence estimator. In Figure~\ref{fig:herakles_architecture} presents how the competence estimator is integrated inside to the model. 

\subsection{$p_g$ numerical calculation}
\label{app:pg}

%At each step $k$ of the high-level policy $\pi^{HL}$, we construct $\tilde{G}_k$ starting from the set of primitive actions and adding to it each goal $g \in G$ with a probability: $$p_g = \max (\mathbb{E}_{\pi^{LL}(.|g), s_k \sim D_{\pi}}[r^{g}| O(s_k)], \varepsilon_k) \, \mathbbm{1}_{nbr\_update\_\pi^{LL}>L} \, \mathbbm{1}_{percentage\_successful\_traj\_buffer>\alpha},$$

%The two indicator functions filter out goals for which insufficient training has been performed or for which too few successful trajectories are available in the training buffer. In such cases, it is reasonable to assume that $\pi^{LL}$ is not yet capable of solving the corresponding goal. While these indicators are not strictly necessary, they introduce useful inductive biases that complement the estimator $\mathbb{E}_{\pi^{LL}(.|g), s_k \sim D_{\pi^{LL}}}[r^{g}| O(s_k)]$.

In Section~\ref{skill_space_generation}, each goal $g \in G$ is included in 
$\tilde{G}_k(o_t)$ via an independent Bernoulli draw with probability 
$p_g = \max\bigl(C_{\theta_k}(o_t, g), \epsilon\bigr)$. In practice, two 
additional filtering conditions are applied. The full inclusion probability 
used in our implementation is:
\begin{equation}
    p_g \;=\; 
    \max\bigl(C_{\theta_k}(o_t, g),\, \epsilon\bigr)
    \;\cdot\; \mathbf{1}_{SR_{\tau}(g) \,>\, \alpha} 
    \;\cdot\; \mathbf{1}_{u^{LL}_g \,>\, L},
    \label{eq:pg_practical}
\end{equation}
where $u^{LL}_g$ denotes the number of low-level updates performed on goal 
$g$ so far, $L$ is a minimum-update threshold, $SR_{\tau}(g)$ is the 
fraction of successful trajectories for $g$ currently stored in the 
buffer $\mathcal{B}^{\text{sg}}_{\text{LL}}$, and $\alpha$ is a 
minimum-success-rate threshold.

The two indicators play distinct roles. The first, $\mathbf{1}_{SR_{\tau}(g) > \alpha}$, 
is a direct consequence of the schedule used to update $\pill$ 
(see Eq.~\ref{eq:ll_update_prob}): $\pill$ is updated on $g$ only once a 
non-trivial fraction of successful trajectories has been collected in the 
buffer, so applying the same condition to $p_g$ ensures that $\tilde{G}_k$ 
is aligned with the goals on which $\pill$ is actually being trained. The second, $\mathbf{1}_{u^{LL}_g > L}$, 
is a simple heuristic: if $\pill$ has not been updated on $g$ a sufficient 
number of times, it is unlikely to solve $g$ regardless of the state, so 
proposing $g$ as a callable option to $\pihl$ would only destabilize 
high-level training. 
Finally, $\epsilon$ retains its role from the main paper: it ensures that 
goals satisfying both indicators keep a minimal inclusion probability, 
sustaining exploration and breaking the self-reinforcing loop between 
$C_{\theta_k}$ and $\tilde{G}_k$.

\subsection{Prompt of the low-level success rate estimator}
\label{app:prompt_ll_sr_estimator}

In this Figure~\ref{fig:example_prompt_ll_sr_estimator}, we give an example of prompts given to the low-level success rate estimator.

\begin{figure}[ht]
\centering
\begin{adjustbox}{max width=\textwidth}
\begin{tcolorbox}[colback=gray!5!white, colframe=gray!75!black, title=Example of prompt of the low-level policy success rate estimator]
\small
\ttfamily
You are playing a Minecraft like game.
Your task is to evaluate your success rate for the goal: \texttt{goal}

Your coordinates: (13,31)

You see:
\begin{itemize}
    \item water 7 steps to your north-west
    \item grass 1 step to your west
    \item stone 3 steps to your north-east
    \item sand 3 steps to your north
    \item table 1 step to your south
\end{itemize}

You face table at your front.

Your inventory:
\begin{itemize}
    \item sapling: 8
    \item stone: 9
    \item coal: 6
    \item wood pickaxe: 3
    \item wood sword: 1
\end{itemize}

You placed table at (39,40)
You placed table at (3,32)
You placed table at (12,34)
You placed table at (13,32)
You placed table at (10,30)
You placed table at (37,18)
You placed table at (40,17)
You placed table at (31,16)

Your success rate is: 
\end{tcolorbox}
\end{adjustbox}
\caption{Example of a prompt given to low-level policy success rate estimator.}
\label{fig:example_prompt_ll_sr_estimator}
\end{figure}

\subsection{Training of the low-level success rate estimator}
\label{app:training_of_LL_sr_estimator}

The low-level success rate estimator predicts the probability of successful skill execution given an observation. It is built upon an LLM backbone, from which we extract the final hidden state of the decoder. This representation is passed through two SiLU-activated hidden layers of size $1024$~\cite{elfwing2017sigmoidweightedlinearunitsneural}, followed by a sigmoid-activated output layer of dimension $1$, yielding a scalar success probability.

During execution, whenever the high-level policy selects a skill, we record a tuple $(\text{obs}_{\text{ll\_sr\_estimator}}, \text{skill}, \text{outcome})$, where \textit{outcome} is a binary indicator of whether the skill was successfully executed by the low-level policy. To augment training data and enhance the estimator’s ability to interpret rich observations, we assume that starting from any of the initial $10\%$ of states visited in a trajectory would not alter the success outcome. Accordingly, we collect $6 = \lfloor 0.1 \times 64 \rfloor$ such tuples per skill execution.

These transitions are aggregated during each high-level data collection cycle (consisting of $2496$ high-level steps; see Appendix~\ref{app:high_level_policy}) and stored in a buffer of size $3$. The estimator is updated every $256$ new transitions using a binary cross-entropy loss and trained for a single update epoch per cycle. These two hyper-parameters were selected based on empirical performance.

\subsection{Composition of the set of skills}
\label{app:i_tilde_composition}

In Section~\ref{skill_space_generation}, we define the probability of sampling a skill $g$ into the admissible skill set $\tilde{G}$ at step $k$ as:
\[
p_g = \max\left(\mathbb{E}_{\pi^{LL}(\cdot|g),\, s_k \sim D_{\pi^{LL}}}\left[r^{g} \mid O(s_k)\right], \varepsilon_{explo}\right),
\]
where $\varepsilon_{explo}$ is an exploration term. We empirically found that $\varepsilon_k$ should reflect the update frequency of the low-level policy corresponding to skill $g$. The rationale is that a frequently updated low-level policy is more likely to have improved, thus increasing the probability that the associated skill will succeed and should be selected.

We therefore define the exploration term as:
\[
\varepsilon_{explo} = \min(\text{update\_frequency\_g},\, 0.1),
\]
where \texttt{update\_frequency\_g} denotes the number of times the low-level policy for skill $g$ has been updated in the past five high-level data collection cycles. This dynamic exploration term outperforms a fixed value, such as the constant $\varepsilon_k = 0.1$, in empirical evaluations.

\newpage
\section{Implementation of MAGELLAN}
\label{app:magellan_implementation}

\subsection{MAGELLAN adaptation}
\label{app:magellan_adaptation}
We reimplemented MAGELLAN from \cite{gaven_magellan2025} to make our autotelic goal sampler prioritize goals with maximum Learning Progress (LP) from $G$. Our implementation slightly differs from the original one to match our framework.

First, the setup in MAGELLAN involved a single goal per episode. As a result, their goal space was the combination of all possible initial states and instructions (i.e., the Multi-Armed Bandit had one arm per initial state-instruction pair). Here, multiple goals can be solved within a single episode. Therefore, whenever a goal has to be sampled, MAGELLAN estimates the LP of each goal from $G$ only for the current state. While this does not change much from a practical point of view, our bandit is now a contextual Multi-Armed Bandit (i.e., the set of arms never changes but the current state conditions the LP estimation).

Second, instead of storing a single entry per goal (i.e. the goal, the state it was sampled from, and its associated outcome) in MAGELLAN's dataset ($\mathcal{D}$ in their paper), we store multiple copies of the same goal-outcome pair with different states seen during the trajectory. In particular, we store the first $N^{HL}*0.1 = 6$ states observed by the high-level policy during the trajectory. This provides richer information for a goal and helps foster generalization.

Then, as a possibly varying number of entries are added to the dataset for each goal (i.e. up to $6$), MAGELLAN's update frequency is now calculated based on the number of entries added instead of goals sampled. We perform this update every $128$ new entries. Moreover, instead of keeping a dataset of fixed size, we use a dataset of varying size storing the entries from the last $3$ data collection phases of the high-level policy (i.e. $2496$ high-level steps, see Appendix~\ref{app:high_level_policy}).  

For the buffer of weights (named $\mathcal{B}$ in MAGELLAN), we only keep the last $3$ success rate estimator copies. Finally, we use an exponential decay for $\epsilon$ with a rate of $3.34$. All the other hyper-parameters not mentioned in the section are kept the same as in the original implementation of MAGELLAN.

\subsection{Prompt of the high-level success rate estimator}
\label{app:training_of_HL_sr_estimator}

We show in \ref{fig:example_prompt_ll_sr_estimator} the prompt given to the LLM when estimating the competence, i.e. the success rate. 

\begin{figure}[ht]
\centering
\begin{adjustbox}{max width=\textwidth}
\begin{tcolorbox}[colback=gray!5!white, colframe=gray!75!black, title=Example Prompt high-level policy success rate estimator]
\small
\ttfamily
You are playing a Minecraft like game.
Your task is to evaluate your success rate for the goal: \texttt{<goal>}

Your coordinates: (24,8)

You see:
\begin{itemize}
    \item grass 1 step to your west
    \item stone 2 steps to your west
    \item path 2 steps to your north-east
    \item coal 2 steps to your south-west
    \item iron 7 steps to your south-west
    \item table 3 steps to your south-east
    \item furnace 1 step to your south
\end{itemize}

You face a furnace at your front.

Your inventory:
\begin{itemize}
    \item sapling: 6
    \item stone: 5
    \item coal: 1
    \item wood pickaxe: 3
    \item wood sword: 1
\end{itemize}

You placed table at (26,36)
You placed table at (16,24)
You placed table at (26,9)
You placed table at (24,0)

You placed plant at (1,0)
You placed plant at (4,0)

You placed stone at (5,17)
You placed stone at (3,17)
You placed stone at (2,18)

You placed furnace at (24,9)

Your success rate is: 
\end{tcolorbox}
\end{adjustbox}
\caption{Example of a prompt given to high-level policy success rate estimator. This prompt is given for each goal in $g \in G$ with <goal> replaced $g$. }
\label{fig:example_prompt_hl_sr_estimator}
\end{figure}

\newpage
\section{Baselines details}
\label{app:baselines_details}

\subsection{FuN}
\label{app:FuN}

We implemented FeUdal Networks (\textsc{FuN}) a hierarchical RL architecture with two components: a Manager and a Worker, both modeled as recurrent networks. The Manager operates at a lower temporal resolution, producing directional goals $g_t \in \mathbb{R}^d$ in a learned latent state space. The Worker receives these goals and produces elementary actions at every time step.

Observations $x_t$ are encoded into latent states $z_t$ via a shared perceptual module. The Manager computes a latent embedding $s_t$, and generates $g_t$ using a dilated LSTM that supports long-range dependencies. The Worker pools recent goals, maps them into a low-dimensional space via a projection $\phi$, and combines them with action embeddings $U_t$ to produce the policy:
\[
\pi_t = \text{SoftMax}(U_t \, \phi\left(\sum_{i=0}^{c-1} g_{t-i}\right))
\]

The Manager is trained using a {\em transition policy gradient} to align $g_t$ with advantageous latent transitions $s_{t+c} - s_t$, while the Worker is trained via intrinsic reward based on goal-following:
\[
r^I_t = \frac{1}{c} \sum_{i=1}^{c} \cos(s_t - s_{t-i}, g_{t-i})
\]
and optimizes a weighted sum of intrinsic and extrinsic rewards.

In the original \textsc{FuN} implementation, only environment observations were passed to the perceptual module, as the evaluated environments were not goal-conditioned. In our version, we concatenate an embedding of the goal to the output of the perceptual module. These goal embeddings are generated using the same language model employed for HERAKLES (Mistral 7B v0.3), enabling the agent to generalize to goals it has not encountered during training.

Table~\ref{tab:fun_hyperparameters} summarizes the key hyper-parameters used in our implementation of \textsc{FuN}.

\begin{table}[h]
\centering
\caption{Hyperparameters used in the \textsc{FuN} implementation.}
\begin{tabular}{lc}
\toprule
\textbf{Parameter}  & \textbf{Value} \\
\midrule
Worker goal horizon & 10 \\
Hidden dimension of Manager & 256 \\
Hidden dimension of Worker & 16 \\
Discount factor for Worker & 0.99 \\
Discount factor for Manager & 0.999 \\
Intrinsic reward coefficient & 0.5 \\
Dilation factor for Manager's LSTM & 10 \\
\bottomrule
\end{tabular}
\label{tab:fun_hyperparameters}
\end{table}

\subsection{POAD training}
\label{app:poad_training}
The \textsc{poad} baseline~\cite{wen_reinforcing2024} corresponds to a simplified version of \herakles in which the hierarchical structure is removed by discarding the low-level policy. In this configuration, all elementary actions are directly selected by the high-level policy. The training procedure strictly follows the methodology and hyper-parameters detailed in Appendix~\ref{app:high_level_policy}.

\subsection{BOSS training}
\label{app:boss_training}
BOSS~\cite{zhang_bootstrap2023} is a method for learning complex long-term tasks by autonomously
growing a skill library.
It proceeds in two phases: (1)~pre-training a language-conditioned skill policy on
a labelled demonstration dataset, and (2)~iteratively bootstrapping that policy into
longer-horizon behaviours. The two phases that composed the method are detailed below with the \textcolor{red}{adaptations} we have made to fit our experimental setup.
 
\paragraph{Phase 1 -- Pre-training a Skill Policy.}
BOSS assumes access to a dataset $\mathcal{D}_L = \{\tau_{z_1}, \tau_{z_2}, \ldots\}$
of trajectories, where each trajectory $\tau_{z_i}$ is annotated with a free-form
language description $z_i$ of a primitive skill (\emph{e.g.}\xspace, ``collect wood'').
A sparse reward function $r$ signals task completion.
A language-conditioned policy $\pi_{LL}^{BOSS}(a \mid s, z)$ and value function $V(s, z)$ are
trained on $\mathcal{D}_L$ using Implicit Q-Learning (IQL)~\cite{kostrikov_offline2022}, yielding a
repertoire of executable primitive skills $\mathcal{Z} = \{z_1, z_2, \ldots\}$.

\textcolor{red}{Adaptations}: In our experimental set up we do not have access to $\mathcal{D}_L$. We replace the primitive skills by elementary actions (\emph{e.g.}\xspace, \textit{move left}, \textit{move right}, \textit{chop tree}, ...) that can always be executed. For modeling $V(s, z)$, we use a value function with a LLM back bone similarly to the one use for $\pihl$ described in Appendix~\ref{app:high_level_policy_architecture}.

\paragraph{Phase 2 -- Skill Bootstrapping.}
Starting from the primitive repertoire, BOSS iterates three steps to compose
increasingly long-horizon skills without additional human supervision.
 
\begin{enumerate}
    \item \textbf{Initial skill sampling.}
          At the start of each episode, a skill $z \in \mathcal{Z}$ is sampled
          proportionally to the pre-trained value function $V(s_1, z)$,
          favouring skills that are likely to succeed from the current state $s_1$.

          \textcolor{red}{Adaptations}: $V(s_1, z)$ is not pretrained in our set up but learned during training on collected trajectories. For elementary actions ( \emph{e.g.} \textit{move left}) that are always successful in one step, we bypass $V$ and assigned a value of $1$ (the true value).
          
    \item \textbf{LLM-guided skill chaining.}
          After a successful skill execution, a large language model (LLM) is prompted with the current
          skill repertoire and the skills executed so far, and proposes the next
          skill in natural language.
          This is repeated $N$ times; the actual next skill is drawn from the
          distribution of LLM-assigned token likelihoods, encouraging diversity.
          The proposal is then mapped back to the closest skill in $\mathcal{Z}$
          via a pre-trained sentence embedding model. $\pi_{LL}^{BOSS}(a \mid s, z)$ is executed to reach the chosen skill, if it is successful step $2$ is repeated $M$ times.

          \textcolor{red}{Adaptations}: the next skills is proposed through constrain decoding among the list of all possible skills in Crafter avoiding the mapping to a pretrained sequence. We use $N=4$ in the experiments. In the original paper they used $M=2$, however, as we start with elementary actions and not pretrained skills, we used $M=4$ to help discovering complex skills. 
          
    \item \textbf{Skill library expansion.}
          At the end of each episode, the collected experience is added to the
          replay buffer with a sparse reward of 1 per completed sub-skill.
          The same LLM is used to generate composite language instructions
          describing multi-step chains (\emph{e.g.}\xspace, ``make wood pickaxe''); these composite instructions and associated trajectories are
          appended to both the replay buffer and the skill repertoire $\mathcal{Z}$
          for further bootstrapping rounds.
          To mitigate catastrophic forgetting, offline data from $\mathcal{D}_L$
          and online data are sampled at equal proportions during policy updates.

          \textcolor{red}{Adaptations}: We use an oracle to relabel the trajectories. Using an LLM leads to highly noisy relabeling in the Crafter environment that impedes the learning of more complex skills. Our oracle relabel the trajectory with the most complex skills reached at the end of it. We store the trajectories in $\mathcal{D}_L$ which is modeled as a queue with a maximum size of $100000$ transitions. We train $V(s_1, z)$ and  $\pi_{LL}^{BOSS}$ $\mathcal{D}_L$ after collecting $48$ trajectories in the environment. 
\end{enumerate}
 
At test time, the LLM is given a prompt similar to that of Appendix~\ref{app:prompts_for_HL}.

%\newpage
\section{Hyper-parameters}
\label{app:hyperparameters}

%\todo{Be more exhaustive}

\begin{table}[h]
\centering
\caption{Hyper-parameters used in \herakles.}
\begin{tabular}{lc}
\toprule
\textbf{Parameter}  & \textbf{Value} \\
\midrule
high-level policy max steps & $64$ \\
low-level policy max steps & $128$ \\
\hline
number of environment in parallel & $48$ \\
number of high level step of data collection for \textsc{poad} & $2496$ \\
$lr_{POAD}$ & $10^{-5}$ \\
$\lambda_{POAD}$ & $0.9$ \\
$\gamma_{POAD}$ & $0.95$\\
entropy coefficient \textsc{poad} & $0.01$ \\
$\beta_{KL}$ & $0.1$ \\
update epoch \textsc{poad} & $4$ \\
\hline
AWR buffer size & $10^5$ \\
$lr_{AWR}$ & $10^{-4}$ \\
\hline
$lr_{\text{low level success rate estimator}}$ & $10^{-4}$ \\
udpate epoch low level success rate estimator & $2$ \\
\bottomrule
\end{tabular}
\label{tab:herakles_hyperparameters}
\end{table}

\section{Compute resources}
\label{app:compute_resources}

To run the various training sessions, we used H100 and V100 GPUs. Training \herakles was done with a cluster consisting of four H100 GPUs for $200$ hours. To train \textsc{POAD}, a cluster comprising four H100s was required for $60$ hours; to train \textsc{BOSS}, a cluster comprising four H100s was required for $100$ hours. Training \textsc{FuN} necessitated $20$ hours on a V100. It is important to note that a significant proportion of the time taken by HERAKLES is accounted for by loading time and buffer operations (approximately $33\%$ of the total time). These operations were not optimised during these experiments

%%%%%%%%%%%%%%%%%%%%%%%%%%%%%%%%%%%%%%%%%%%%%%%%%%%%%%%%%%%%

\newpage

\section*{NeurIPS Paper Checklist}

\begin{enumerate}

\item {\bf Claims}
    \item[] Question: Do the main claims made in the abstract and introduction accurately reflect the paper's contributions and scope?
    \item[] Answer: \answerYes{} % Replace by \answerYes{}, \answerNo{}, or \answerNA{}.
    \item[] Justification:  Section \ref{experiments} provides evidence supporting the claims made in the abstract.

\item {\bf Limitations}
    \item[] Question: Does the paper discuss the limitations of the work performed by the authors?
    \item[] Answer: \answerYes{} % Replace by \answerYes{}, \answerNo{}, or \answerNA{}.
    \item[] Justification: Justification: Section \ref{sec:conclusion} discusses the limitations.

\item {\bf Theory assumptions and proofs}
    \item[] Question: For each theoretical result, does the paper provide the full set of assumptions and a complete (and correct) proof?
    \item[] Answer: \answerNA{} % Replace by \answerYes{}, \answerNo{}, or \answerNA{}.
    \item[] Justification: Justification: The paper does not introduce new theoretical results.

    \item {\bf Experimental result reproducibility}
    \item[] Question: Does the paper fully disclose all the information needed to reproduce the main experimental results of the paper to the extent that it affects the main claims and/or conclusions of the paper (regardless of whether the code and data are provided or not)?
    \item[] Answer: \answerYes{} % Replace by \answerYes{}, \answerNo{}, or \answerNA{}.
    \item[] Justification: We provide detailed descriptions of our experimental settings, model architectures, and training procedures in the paper and supplementary materials. We also release our codebase (see the Appendix~\ref{app:herakles_implementation}), including all configuration files needed to reproduce our experiments.

\item {\bf Open access to data and code}
    \item[] Question: Does the paper provide open access to the data and code, with sufficient instructions to faithfully reproduce the main experimental results, as described in supplemental material?
    \item[] Answer: \answerYes{} % Replace by \answerYes{}, \answerNo{}, or \answerNA{}.
    \item[] Justification: We provide detailed descriptions of our experimental settings, model architectures, and training procedures in the paper and Appendix~\ref{app:hierarchical_agent}.We also release our codebase, including all configuration files needed to reproduce our experimentsAppendix~\ref{app:herakles_implementation}.

\item {\bf Experimental setting/details}
    \item[] Question: Does the paper specify all the training and test details (e.g., data splits, hyperparameters, how they were chosen, type of optimizer) necessary to understand the results?
    \item[] Answer: \answerYes{} % Replace by \answerYes{}, \answerNo{}, or \answerNA{}.
    \item[] Justification: All details are specified in Section \ref{experiments} and the Appendices~\ref{app:hierarchical_agent}~\ref{app:hyperparameters}.

\item {\bf Experiment statistical significance}
    \item[] Question: Does the paper report error bars suitably and correctly defined or other appropriate information about the statistical significance of the experiments?
    \item[] Answer: \answerYes{} % Replace by \answerYes{}, \answerNo{}, or \answerNA{}.
    \item[] Justification: Experiments have been done using $5$ seeds and standard deviation is reported in the figures and tables in Section \ref{experiments} and Appendix~\ref{app:complementary_results}.

\item {\bf Experiments compute resources}
    \item[] Question: For each experiment, does the paper provide sufficient information on the computer resources (type of compute workers, memory, time of execution) needed to reproduce the experiments?
    \item[] Answer: \answerYes{} % Replace by \answerYes{}, \answerNo{}, or \answerNA{}.
    \item[] Justification: Information on the compute resources are provided in the Appendix~\ref{app:compute_resources}.
    
\item {\bf Code of ethics}
    \item[] Question: Does the research conducted in the paper conform, in every respect, with the NeurIPS Code of Ethics \url{https://neurips.cc/public/EthicsGuidelines}?
    \item[] Answer: \answerYes{} % Replace by \answerYes{}, \answerNo{}, or \answerNA{}.
    \item[] Justification: Our research adheres to the code of ethics by using only synthetic, non-personal data from a simulated environment, posing no risk to human subjects, privacy, or societal harm, and we commit to open-sourcing our code and models with proper documentation to ensure reproducibility and responsible use.

\item {\bf Broader impacts}
    \item[] Question: Does the paper discuss both potential positive societal impacts and negative societal impacts of the work performed?
    \item[] Answer: \answerNo{}
    \item[] Justification: While our results demonstrate significant improvement in LLM agents learning increasingly complex skills, our experiments were limited to small-scale LLMs and well-controlled testbeds. Therefore, we caution against generalizing these findings to real-world open-ended learning scenarios.
    
\item {\bf Safeguards}
    \item[] Question: Does the paper describe safeguards that have been put in place for responsible release of data or models that have a high risk for misuse (e.g., pre-trained language models, image generators, or scraped datasets)?
    \item[] Answer: \answerNA{} % Replace by \answerYes{}, \answerNo{}, or \answerNA{}.
    \item[] Justification: The paper does not pose any such risks.

\item {\bf Licenses for existing assets}
    \item[] Question: Are the creators or original owners of assets (e.g., code, data, models), used in the paper, properly credited and are the license and terms of use explicitly mentioned and properly respected?
    \item[] Answer: \answerYes{} % Replace by \answerYes{}, \answerNo{}, or \answerNA{}.
    \item[] Justification: We cite and comply with the licenses of the LLMs used in the experiments in Section \ref{experiments}.

\item {\bf New assets}
    \item[] Question: Are new assets introduced in the paper well documented and is the documentation provided alongside the assets?
    \item[] Answer: \answerYes{} % Replace by \answerYes{}, \answerNo{}, or \answerNA{}.
    \item[] Justification: The codebase and implementation details are shared alongside the paper.

\item {\bf Crowdsourcing and research with human subjects}
    \item[] Question: For crowdsourcing experiments and research with human subjects, does the paper include the full text of instructions given to participants and screenshots, if applicable, as well as details about compensation (if any)? 
    \item[] Answer: \answerNA{} % Replace by \answerYes{}, \answerNo{}, or \answerNA{}.
    \item[] Justification: The paper does not involve experiments with human participants.

\item {\bf Institutional review board (IRB) approvals or equivalent for research with human subjects}
    \item[] Question: Does the paper describe potential risks incurred by study participants, whether such risks were disclosed to the subjects, and whether Institutional Review Board (IRB) approvals (or an equivalent approval/review based on the requirements of your country or institution) were obtained?
    \item[] Answer: \answerNA{} % Replace by \answerYes{}, \answerNo{}, or \answerNA{}.
    \item[] Justification: The paper does not involve experiments with human participants.

\item {\bf Declaration of LLM usage}
    \item[] Question: Does the paper describe the usage of LLMs if it is an important, original, or non-standard component of the core methods in this research? Note that if the LLM is used only for writing, editing, or formatting purposes and does \emph{not} impact the core methodology, scientific rigor, or originality of the research, declaration is not required.
    %this research? 
    \item[] Answer: \answerNA{} % Replace by \answerYes{}, \answerNo{}, or \answerNA{}.
    \item[] Justification: The core method development in this research does not involve LLMs as any important, original, or non-standard components.

\end{enumerate}

\end{document}